\documentclass{article} 
\usepackage[final]{colm2026_conference}

\usepackage{microtype}
\usepackage{hyperref}
\usepackage{url}
\usepackage{booktabs}
\usepackage{xcolor}
\usepackage{array}
\usepackage{amsmath}
\usepackage{amssymb}
\usepackage{graphicx}
\usepackage{wrapfig}
\usepackage{multirow}
\usepackage{lineno}
\usepackage{tabularx}
\usepackage{float}
\usepackage[most]{tcolorbox}

\definecolor{darkblue}{rgb}{0, 0, 0.5}
\hypersetup{colorlinks=true, citecolor=darkblue, linkcolor=darkblue, urlcolor=darkblue}

\title{Learning to Refer from Estimated Listener Gaze}

\author{Téa Wright \& Alane Suhr \\
Department of Computer Science\\
University of California, Berkeley\\
Berkeley, CA 94704, USA \\
\texttt{\{teaywright, suhr\}@berkeley.edu} \\
}

\begin{document}

\ifcolmsubmission
\linenumbers
\fi

\maketitle

\begin{abstract}
We propose to finetune vision-language models to generate more pragmatically optimal referring expressions by transforming observations of incremental listener comprehension, in the form of gaze scanpaths, into learning signals. 
During training, referring expressions are sampled from the speaker policy being optimized, conditioned on images and target referents; then, a neural listener estimating human gaze behavior maps from images and sampled referring expressions to scanpaths, each represented by a sequence of fixations, with each fixation corresponding to a word in the referring expression.
We experiment with several approaches to convert fixation sequences and target referents into token- and sequence-level rewards, which are used to optimize policy parameters.
Through evaluation with human listeners, we find that speaker policies trained with gaze-estimating listeners result in significantly more pragmatically-optimal references than base models, reducing sequence length from 15.4 down to 4.0 words while increasing referential success from 75.2 up to 80.0\%. 
Our work demonstrates a promising opportunity for learning to generate utterances through language-based interaction, not only from the explicit signal of communicative success, but also from implicitly-available observations of a listener's process of comprehension. \footnote{Code and model weights are available at: \url{https://github.com/Berkeley-NLP/reg-from-gaze}}


\end{abstract}
\section{Introduction} 



Language-based interaction is collaborative.
People collaborate through language use to come to a shared understanding of the world, and they do so efficiently by minimizing the total combined effort necessary to reach mutual understanding~\citep{clark1986referring}.
In natural human interaction, we minimize joint effort through communication signals beyond the textual content of our utterances, including backchanneling, paralanguage, and gaze. 
Reference games are an ideal testbed to understand and measure these phenomena.
Here, a speaker and a listener collaborate to establish joint attention to a part of their shared environment~\citep{dale1995computational}.
The speaker's task, known as referring expression generation (REG), is to produce an utterance that uniquely identifies the region of interest to their listener. The listener's task is to identify the target based on the speaker's description, called referring expression comprehension (REC). 
In reference games and elsewhere, people design their utterances by taking into account the shared interaction situation and their knowledge of their audience so that their contributions to the interaction are as informative as necessary, but not overly verbose~\citep{grice1975logic,kapur2026wordssaylessdecoupling}.

On the other hand, neural language models do not exhibit this level of pragmatic optimality. 
\cite{Ma2025-pd} finds that referring expressions generated by modern vision-language models (VLMs) in reference games differ greatly from human-written references; they tend to include unnecessary information and often fail to uniquely identify referents by taking into account the interaction context.
Recent work has approached the problem of improving referential success through learning from interaction, where a VLM policy is rewarded for successfully describing an intended referent to a listener~\citep{gul2024cogen,tang-etal-2024-grounding}.
These policies are trained only from the outcome of reference games.
Instead, \citet{jara2022social} argue that REG is best understood as a pragmatic effort to guide the listener's visual search process, observations of which are used by speakers to check listener understanding during language production~\citep{clark2004speaking}.

\begin{figure}[t]
\begin{center}
\centering
   \includegraphics[width=\linewidth,clip,trim={30 350 100 100}]{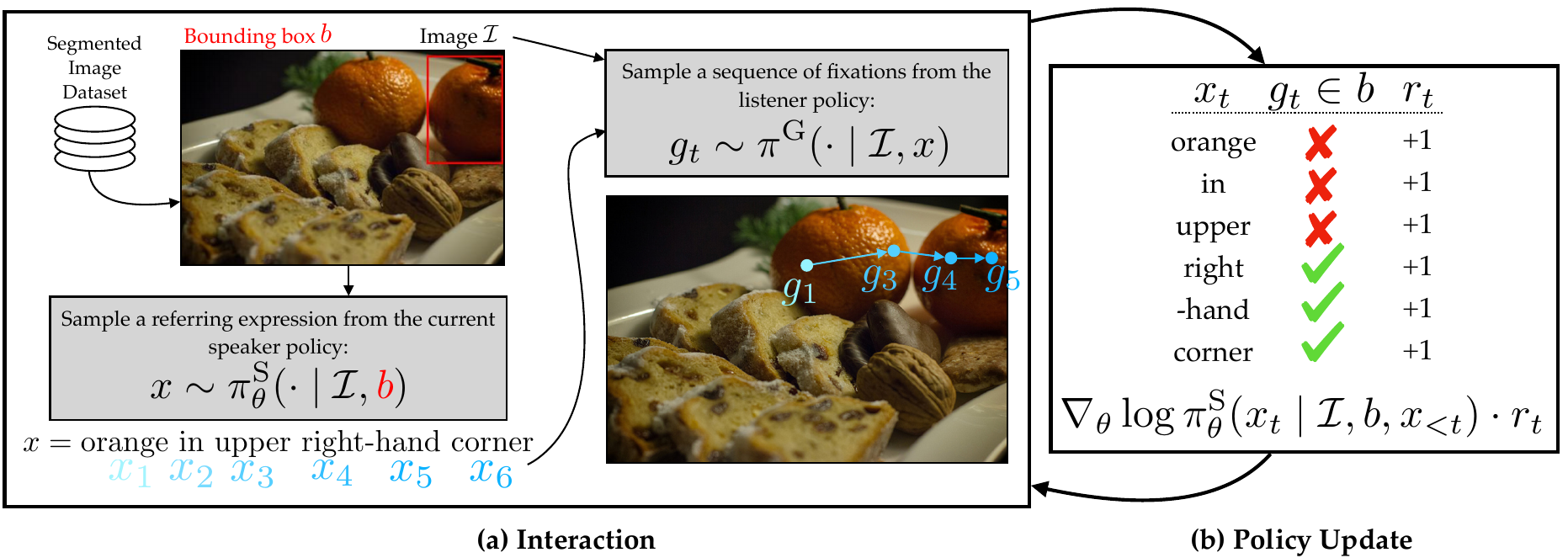}
\end{center}
\caption{Learning to refer from interaction with incremental listeners. During training, we iterate between (a) deploying a speaker policy $\pi^\text{S}_\theta$ in a reference game with a listener policy $\pi^\text{G}$, which is trained to estimate gaze behavior of human listeners, and (b) updating the speaker policy by transforming observations of listener gaze behavior to rewards.
}
\label{fig:fig1}
\end{figure}

We use observations of the process of visual search during reference comprehension as a learning signal for improving VLMs as speakers in reference games.
Gaze provides a quantifiable signal that allows us to directly relate a speaker's language to a listener's visual search process; we can test exactly which tokens meaningfully contribute to the listener's understanding.
To this end, we design a learning setup where VLMs are finetuned as speakers in interaction with gaze-estimating listeners. 
First, we train a VLM to estimate listener gaze. 
Then, during training of the VLM speaker, we iterate between deploying it in reference games with the gaze-estimating listener (Figure~\ref{fig:fig1}a), and optimizing the speaker policy using rewards derived from listener gaze estimates (Figure~\ref{fig:fig1}b).

We experiment with a variety of reward functions for converting gaze estimates to learning signals, and compare our approach to learning from non-gaze-estimating listeners. 
Our results show that using gaze-estimating listeners during training can both boost referential success in reference games with human listeners from 75.2 up to 80.0\%, while also reducing the verbosity of such references from 15.4 down to 4.0 words on average.
Our results also reveal a clear tradeoff in current VLMs between accuracy and verbosity: while speakers trained with non-gaze-estimating listeners achieve even higher referential success (up to 86\%), this comes at a cost of generating increasingly longer references, up to an average of 24.4 words.
Overall, we find that learning from estimated observations of the \textit{process} of reference comprehension, rather than the \textit{outcome} of a reference game alone, results in speaker policies that generate much more pragmatically optimal references.




\section{Technical overview}

\noindent\textbf{Problem statement} Let $x = \langle x_1, \dots, x_T \rangle \in \mathcal{V}^+$ be a natural language referring expression comprised of tokens from a vocabulary $\mathcal{V}$, $\mathcal{I} \in \{0,\dots,255\}^{M \times N \times 3}$ an image of size $M$ by $N$, and $b = (b_x, b_y, b_w, b_h)\in \mathbb{Z}^4$ a bounding box circumscribing a target referent.\footnote{$0 \leq b_x \leq M$ and $0 \leq b_y \leq N$ represent the anchor point of the bounding box at its top left corner; $0 < b_w \leq M-b_x$ and $0 < b_h \leq N - b_y$ represent the width and height of the bounding box.}
We study the problem of referring expression generation, where a speaker policy $\pi^\text{S}_\theta$ maps from an image $\mathcal{I}$ and target bounding box $b$ to a referring expression $x$.
The task of referring expression comprehension (REC) is executed by a listener policy $\pi^\text{REC}$, which maps from the image $\mathcal{I}$ and the generated referring expression $x$ to a coordinate $r = (r_x, r_y) \in \{0, \dots, M\} \times \{0, \dots, N\}$.
Communicative success is achieved if the listener correctly identifies the target referent circumscribed by $b$; i.e., $r \in b$.\footnote{We say a coordinate $(x, y) \in b = (b_x, b_y, b_w, b_h)$ (a ``hit'') if $b_x \leq x \leq b_x + b_w$ and $b_y \leq y \leq b_y + b_h$.} 
Success requires that $x$ describes the target referent accurately and unambiguously with respect to the entire image.
For every image and target referent, there are a large number of possible referring expressions that satisfy these criteria; one open challenge, then, is achieving referential efficiency by identifying the shortest such referring expression.
This is the focus of our study.
We extend this classical formulation of referring expression comprehension, where a listener maps from an entire referring expression to a single candidate coordinate, to an \textit{incremental} comprehension task. 
Let a scanpath $g = \langle g_1, \dots, g_T \rangle$ be a sequence of gaze fixations, with each $g_i$ taking on either a null value $\emptyset$, or a coordinate value $\left(g^{(i)}_x, g^{(i)}_y\right) \in \{0, \dots, M\} \times \{0, \dots, N\}$.
Here, a listener policy $\pi^\text{G}$ maps from an image $\mathcal{I}$, (possibly partial) referring expression $x$ of length $t \leq T$, and a scanpath prefix $g$ of length $t-1$, to the next fixation $g_t$.
Reference comprehension is performed iteratively; each fixation $g_t \sim \pi^{\text{G}}(\cdot \mid \mathcal{I}, x_{1:t}, g_{1:{t-1}})$ corresponds to an input word $x_t$.





\noindent\textbf{Approach} Our goal is to learn a speaker policy $\pi^\text{S}_\theta$ that generates referring expressions which are both \textit{effective} and \textit{efficient}.
$\pi^\text{S}$ is instantiated as a large VLM, with initial parameters $\theta_0$ equivalent to the pretrained Molmo-7B-D model~\citep{deitke2024molmopixmoopenweights}.
$\pi^\text{S}_{\theta_0}$ is neither efficient nor effective as a speaker; references generated from this policy are overly verbose and result in low communicative success.
We finetune this policy through reinforcement learning to maximize reward (Section~\ref{sec:training}).
During training, we simulate REC with a listener model $\pi^\text{G}_\phi$ that we have trained to estimate human gaze scanpaths (Section~\ref{sec:gaze_pred}).
Speaker reward is a function not only of final communicative success, but also of the incremental process of reference comprehension as estimated by the listener model $g \sim \pi^\text{G}_\phi(\cdot \mid \mathcal{I}, x)$.


\section{Related works}

Referring expression generation (REG) is often framed as an optimization problem where a speaker must identify the lowest-cost (e.g., shortest) utterance which uniquely distinguishes a target referent in the context of distractors.
While this optimization is typically modeled at a whole-utterance level, e.g., via the rational speech acts (RSA) model~\citep{frank2012predicting,goodman2016pragmatic}, \cite{cohn-gordon-etal-2018-pragmatically} achieve improved communicative success with LSTM-based speakers using incremental character-level RSA.
Modern VLMs are typically able to accurately describe target referents in isolation, but they very often generate pragmatically suboptimal utterances that are unnecessarily verbose~\citep{kapur2026wordssaylessdecoupling} or ambiguous with respect to the entire image, resulting in both low communicative success and inefficient interaction~\citep{Ma2025-pd}.

Recent work finetunes existing VLMs to produce more pragmatically optimal references by learning from communicative success through interaction with listeners~\citep{willemsen-skantze-2024-referring-expression,tang-etal-2024-grounding,gul2024cogen}.
During learning, references are sampled from the speaker policy (i.e., REG), then sent to a listener, who selects the point or region of the image that the sampled reference most likely refers to (i.e., REC).
References that result in communicative success are kept as positive examples, while references for which the listener failed to identify the target are kept as negative examples. 
This prior work varies in how positive and negative examples are converted to learning signals; for example, \cite{gul2024cogen} cast learning as a contextual bandit problem, assigning positive rewards to positive examples and negative rewards to negative ones.
However, in all existing work, the learning signal is sparse: each reference is paired with only one piece of feedback, dependent on the final \textit{outcome} of a reference game.
Here, we use observations of the \textit{process} of  comprehension as a learning signal, as evidenced by the trajectory of listener gaze during reference comprehension.

In machine learning, human gaze data has been modeled and used as a behavior of interest both as a communication device and learning signal, particularly in natural language processing (NLP) and human-robot interaction (HRI). Gaze and attention data have also been used to guide language generation in descriptive visual tasks like image captioning, video captioning, and video summarization \citep{Yu2017-mj, Chen2018-em, xu2015gaze}. 
In HRI, gaze of human operators in jointly embodied environments with robots has been studied as communicative signal for establishing joint attention~\citep{huang2016anticipatory,ruppel2025gaze}, but embodied gaze has not been explored as a signal from which general language-using agents can learn.
In NLP, human gaze patterns during reading of written text have been used as a learning signal to improve performance on text-related tasks ~\citep{Mishra_Kanojia_Bhattacharyya_2016,barrett2020sequence,Kiegeland_undated-vq,papadopoulos-alignment,bondar-etal-2025-aleyegnment, lopezcardona2025seeingeyeaihuman}.
However, this work is limited to non-embodied settings, where gaze provides a signal more about human processing of the text itself rather than its grounding to the world.

Existing work on gaze estimation focuses mainly on free-viewing tasks \citep{Kummerer2022-gf, Kummerer2023-gc,yin2024lg} and visual search \citep{Hu2021-ap, Cartella_2025_ICCV,cartella2025modeling, mondal2025gaze} conditioned on simple single-word references. Some recent works study gaze estimation during more complex language understanding tasks, such as visual question answering~\citep{chen2021predicting}. 
Most similar to our approach, \citet{Mondal2024-ul} study estimating human gaze during REC by collecting scanpaths over RefCOCO \citep{kazemzadeh-etal-2014-referitgame} and training a corresponding gaze-prediction model with a single-stream VLM architecture. 



\section{Estimating listener gaze}
\label{sec:gaze_pred}
We train an incremental listener policy, which we refer to as Molmo-REC-Gaze, to estimate human gaze scanpaths during reference comprehension.
We initialize the policy's parameters as Molmo-7B-D~\citep{deitke2024molmopixmoopenweights} and finetune it using examples adapted from the RefCOCO-Gaze dataset~\citep{Mondal2024-ul}.
This approach results in gaze scanpath estimation superior to previous approaches such as ART~\citep{Mondal2024-ul}.

\subsection{Data} 
RefCOCO~\citep{kazemzadeh-etal-2014-referitgame} is a popular corpus used for the study of multimodal reference.
\cite{Mondal2024-ul} introduce RefCOCO-Gaze, which labels a subset of examples from RefCOCO with scanpaths of human participants' gaze during reference comprehension.
We adapt this dataset to train a neural model that estimates the gaze scanpath of a human listener for arbitrary images and referring expressions. 
To acquire each scanpath, they display the image (without the bounding box annotation) to the participant, and instruct the participant to focus their attention to the center of the image.
The human-written referring expression is then verbalized 
with word onset timestamps recorded.
During reference comprehension, an infrared video-based eye-tracker records participant scanpaths, each of which comprises a sequence of fixations paired with timestamps.
Each scanpath in RefCOCO-Gaze demonstrates successful referential comprehension.

\noindent\textbf{Word and fixation alignment} To create demonstration data for our incremental listener policy, we must pair each word in the referring expression to a single fixation. 
Word onset and fixation timestamps included in RefCOCO-Gaze allow us to identify, for each fixation, which word was most recently heard by the participant.
We apply a 200ms delay to the timestamp of each word to account for auditory processing time~\citep{kirchner2006ultra}.
There may be multiple fixations per word; when this happens, we select the final fixation before the subsequent word onset.
In some cases, there are no corresponding fixations for a word, for example if the word is very short, resulting in a null point $\emptyset$.
We convert the RefCOCO-Gaze training set, which contains $D_T = $ 16,982 scanpaths across 1,799 RefCOCO examples, into a set of demonstrations $\mathcal{D}^\text{train}_\text{G}=\left\{x^{(i)}, g^{(i)}\right\}_{i=1}^{D_T}$, where $\left|x^{(i)}\right|=\left|g^{(i)}\right|$.
For each example, $x_0 = \texttt{BOS}$, $g_0$ indicates listener gaze before the audio is played, and $x_T = \texttt{EOS}$.
The final fixation $g_T \in b$ is always in the target bounding box.
We also convert the validation set into a dataset $\mathcal{D}_\text{G}^\text{val}$ comprising 869 scanpaths across 92 RefCOCO examples.


\subsection{Finetuning for gaze estimation}
Most existing VLMs are trained to perform REC by mapping from an image and a referring expression to bounding box coordinates.
The Molmo family of VLMs~\citep{deitke2024molmopixmoopenweights} are instead trained to perform REC by mapping form an image and referring expression to a set of point coordinates, which makes it an ideal initialization for estimating gaze.
We initialize our gaze listener policy as Molmo-7B-D and optimize it to minimize the token-level loss of gaze sequences included in $\mathcal{D}_\text{G}$, i.e., we approximate \begin{small}$$\phi = \arg \max_{\phi'}\left(-\sum_{i=1}^{D_T}\sum_{t=0}^{\left|x^{(i)}\right|}\log \pi^\text{G}_{\phi'}\left(g^{(i)}_t\mid x^{(i)}_{\leq t}, g^{(i)}_{<t}\right) \right) \; \; .$$\end{small}
Words in $x$ may be tokenized  into multiple subwords; we label each token with the same fixation label.
We train for 4 epochs and select the best checkpoint based on loss on the validation set $\mathcal{D}_\text{G}^\text{val}$. 
We refer to the final trained policy, which selects $g_t \approx \arg \max \pi^\text{G}_\phi(\cdot \mid x_{\leq t}, g_{<t})$, as Molmo-REC-Gaze.
Figure ~\ref{fig:gaze-est-ex} compares scanpaths from human participants from the validation set $\mathcal{D}_\text{G}^\text{val}$ with scanpaths estimated by Molmo-REC-Gaze.

\noindent\textbf{Results} \cite{Mondal2024-ul} release ART, a VLM trained on RefCOCO-Gaze to estimate human listener gaze, which we evaluate as a baseline to Molmo-REC-Gaze.
We report two metrics: (a) path distance, based on dynamic time warping \citep{senin2008dynamic},  compares a human gaze scanpath with an estimated scanpath; and (b) REC accuracy, the frequency at which an estimated scanpath ends in the target bounding box (REC accuracy).
On the validation set $\mathcal{D}^\text{val}_\text{G}$,\footnote{The test set of RefCOCO-Gaze is not publicly available.} Molmo-REC-Gaze significantly outperforms ART, reducing path distance to the closest human scanpath from 55.9 to 39.9 and increasing REC accuracy from 66.7 to 84.4\%. See Appendix~\ref{appx:gazepred} for implementation details, formatting changes, and qualitative examples of gaze prediction. 

\section{Learning from estimated gaze}
\label{sec:training}

Our overall learning setup follows existing work on continual learning from interaction~\citep{kojima-etal-2021-continual,suhrartzi2023,gul2024cogen}, where training iterates between deploying the policy being optimized in interaction with another language-using agent (Figure~\ref{fig:fig1}a), and optimizing the policy based on observations of the interaction (Figure~\ref{fig:fig1}b).
In our work, the policy being optimized is a VLM $\pi^\text{S}_\theta$ that maps from images and bounding boxes to a referring expression (i.e., REG), and the listener agent is Molmo-REC-Gaze, which maps from images and referring expressions to an estimated word-aligned scanpath.

\noindent\textbf{Deploying the speaker policy} We assume access to a dataset $\mathcal{D}_\text{I}$ of images, where each image is annotated with object bounding boxes.
In practice, $\mathcal{D}_\text{I}$ is the validation split of COCO-2014 \citep{lin2015microsoftcococommonobjects}.\footnote{We use the validation set to avoid overlap with images on which Molmo-REC-Gaze was trained, as RefCOCO images (thus, RefCOCO-Gaze images) were taken from the COCO-2014 training set.}
We randomly split $\mathcal{D}_\text{I}$ into training $\mathcal{D}^\text{train}_\text{I}$ (95\%, 38,131 images) and validation $\mathcal{D}^\text{val}_\text{I}$ (5\%, 2,006) splits.
During each training interaction, we sample an image $\mathcal{I}$ from $\mathcal{D}^\text{train}_\text{I}$, then sample a bounding box $b$ whose size is between 1 and 30\% of the image's area.
This removes target referents that are trivially easy to describe (too large), and ones which are too difficult to perceive (too small). We then sample a referring expression from the current speaker policy; i.e., $x \sim^\tau \pi^{\text{S}}_\theta (\cdot \mid \mathcal{I},b)$. 
Bounding boxes are annotated on the image as a transparent rectangle with a 3-pixel red border.
We subsequently estimate a gaze scanpath $g$ from Molmo-REC-Gaze, conditioned on the image $\mathcal{I}$ and sampled referring expression $x$.
To avoid a listener spuriously identifying the target referent, we set the initial pre-reference fixation $g_0$ to a random coordinate outside of the bounding box $b$.
Each interaction results in an example $(\mathcal{I}, b, x, g)$ such that $|x| = |g|$.

\noindent\textbf{Optimizing policy from interaction observations} We experiment with four reward functions that transform our interaction observation $(\mathcal{I}, b, x, g)$ into a learning signal. 
Each reward function assigns a reward $r_t$ to each of the generated tokens $x_{t\geq1}$, taking into account whether fixations in the estimated gaze scanpath are in the bounding box.

\begin{wrapfigure}{r}{3.2cm}\vspace{-1\baselineskip}
\begin{small}\boxed{r_t^{\text{SeqLPHit}} = \mathbb{I}[g_{T'} \in b]}\end{small}
\vspace{-1\baselineskip}\end{wrapfigure}
\textit{Sequence-level last-point hit} (\textssc{SeqLPHit}). Each token $x_t$ receives a positive reward only if the final non-null fixation $g_{T'}$ is in the bounding box, where $T' = \max \{ t \leq T : g_t \neq \emptyset \}$. This function rewards all tokens only if the listener's gaze settles on the intended target at the end of the reference.

\begin{wrapfigure}{r}{5.7cm}\vspace{-1\baselineskip}
\begin{small}\boxed{
r_t^{\text{SeqAnyHit}} = \mathbb{I}[\exists k \in \{1, \dots, T\} : g_k \in b]}\end{small}
\vspace{-1\baselineskip}\end{wrapfigure}
\textit{Sequence-level any hit} (\textsc{SeqAnyHit}). Each token $x_t$ receives a positive reward if any fixations in the gaze scanpath are within the bounding box $b$. We use this reward to identify if the speaker provided sufficient information to ground the referent at any point during the utterance.

\begin{wrapfigure}{r}{5.7cm}\vspace{-1\baselineskip}
\begin{small}\boxed{r_t^{\text{BeforeFirstHit}} = \begin{cases} 1 & \text{if } 1 \leq t \leq \tau \\ 1 & \text{if } t = T \text{ and } \tau > 0\\ 0 & \text{otherwise} \end{cases}}\end{small}\vspace{-1\baselineskip}\end{wrapfigure}
 \textit{Before first hit} (\textsc{BeforeFirstHit}). If the interaction demonstrates communicative success, we identify the timestep of the first hit $\tau = \min\{k : g_k \in b\}$; otherwise, we set $\tau=0$. Then, we assign a positive reward to $x_\tau$ and all tokens in its prefix $x_{1\leq t<\tau}$. For successful interactions, we assign a positive reward to the \texttt{EOS} token $x_T$. The goal of this function is to only reinforce the tokens necessary to reach the earliest point of understanding.

\begin{wrapfigure}{r}{4.5cm}\vspace{-1\baselineskip}
\begin{small}\boxed{r_t^{\text{Shaping}} =\frac{d(g_{t-1}, b) - d(g_t, b)}{d(g_0, b)}}\end{small}
\vspace{-1\baselineskip}\end{wrapfigure}
\textit{Distance-based shaping} (\textsc{Shaping}). We experiment with reward shaping~\citep{ng1999} using a potential function based on the distance of each fixation $g_t$ from the bounding box $b$ relative to the initial fixation $g_0$. We define the distance metric $d(g, b)$ to be the Euclidean distance between the fixation point $g$ and the closest point in the bounding box $b$. Thus, the reward for each token $x_t$ is the change in relative distance from the fixation associated with the previous token. This rewards tokens that progressively guide the listener's gaze closer to the target, even before a "hit" occurs.

After assigning rewards, we use REINFORCE~\citep{sutton1999} to compute and apply gradients to policy parameters:
 \begin{small}$$\nabla_\theta \mathcal{J}(\theta) = \mathbb{E}_{(\mathcal{I}, b, x, t)} (r_t-\underbrace{\beta \text{KL}\left(\pi^\text{S}_{\theta}, \mathcal{I}, b, x, t \right)}_{\text{KL penalty}}) \underbrace{\nabla_\theta (\log \pi^\text{S}_{\theta}\left(x_t\mid\mathcal{I}, b, x_{<t})\right)}_\text{Policy gradient} + \underbrace{\alpha \nabla_\theta H \left(\pi^\text{S}_{\theta}(x_t\mid\mathcal{I}, b, x_{<t})\right)}_{\text{Entropy bonus}} \; \;,$$\end{small}
where $\text{KL}\left(\pi^\text{S}_{\theta}, \mathcal{I}, b, x, t\right) = D_\text{KL}\left(\pi^\text{S}_{\theta}(x_t\mid\mathcal{I}, b, x_{<t}) \mid\mid \pi^\text{S}_{\theta_0}(x_t\mid\mathcal{I}, b, x_{<t})\right)$ is the Kullback-Leibler distance between the current policy with parameters $\theta$ and the initial policy $\theta_0$. 

\section{Experimental Setup}

Our experiments focus on fine-tuning a Molmo~\citep{deitke2024molmopixmoopenweights} model as a speaker.
Evaluation targets two metrics: reference accuracy and efficiency.
We compare our approach with learning from interaction with a non-gaze listener model, referred to as Molmo-REC.

\noindent\textbf{Speaker model training} We initialize the speaker policy $\pi^\text{S}$ with $\theta_0$ as Molmo-7B-D-0924~\citep{deitke2024molmopixmoopenweights}.
When sampling, we use temperature $\tau =$ 0.8, and randomly select a prompt from the brief prompts in \cite{Ma2025-pd} for each episode. We linearly normalize rewards within-batch to span $[-1, 1]$. 
We use KL penalty coefficient of $\beta=$ 0.02 and an entropy regularization coefficient of $\alpha=$ 0.01. In all experiments, we finetune for 1,000 optimizer steps using an effective batch size of 64 and learning rate of 1e-5. We apply LoRA finetuning \citep{hu2021loralowrankadaptationlarge} (rank 8; alpha 16) and checkpoint parameters every 2,000 episodes, selecting the best checkpoint for each run using the reward on the held-out validation set $\mathcal{D}^\text{val}_\text{I}$.

\noindent\textbf{System variations} 
In addition to the models trained in the main experiments using the four different reward functions (\textsc{SeqLPHit}, \textsc{SeqAnyHit}, \textsc{BeforeFirstHit}, and \textsc{Shaping}; see Section~\ref{sec:training}), we compare with (a) the pre-trained Molmo model without any fine-tuning (\textsc{Molmo}), which serves as a baseline, and (b) human-written references (\textsc{Human}), which serve as an upper bound. We also experiment with two additional interaction settings during learning, where the listener is a pre-trained Molmo-7B-D model.
As with our proposed models, we compute a scalar reward for each token $x_t$ based on observations of listener comprehension, and apply REINFORCE.
Interaction data and rewards for examples $(\mathcal{I}, b, x)$ are computed as follows: 

\begin{wrapfigure}{r}{3.8cm}\vspace{-\baselineskip}
\begin{small}\boxed{r_t^{\text{REC-Success}} = \mathbb{I}[g_{\text{REC}} \in b]}\end{small}
\vspace{-2\baselineskip}\end{wrapfigure}
\textit{Communicative success} (\textsc{REC-Success}). We condition a Molmo model on image $\mathcal{I}$ and referring expression $x$ and instruct it to point at the object being referred to, resulting in a single coordinate $g_\text{REC}$. 
This method removes any signals of incremental comprehension.

\textit{Iterative prompting} (\textssc{REC-Iter}). For each prefix in a referring expression $x_{\leq t}$, we condition a Molmo model on image $\mathcal{I}$ and prefix $x_{\leq t}$, and instruct it to point at the object being referred to. This results in a sequence of coordinates  $\tilde{g}$.
This allows us to use the same reward functions we use with $\pi^G$ in an effort to disentangle effects of the listener policy and reward design. Policies using this listener are indicated by the prefix \textsc{REC-}.

\begin{table}[t]
\centering
\small
\begin{tabularx}{\linewidth}{lccccc}
\toprule
\textbf{Speaker} & \textbf{$d_{\pi^s}$ (Human)} $\downarrow$ & \textbf{$d_{\pi^s}$ (Qwen)} $\downarrow$ & \textbf{Acc.} $\uparrow$ & \textbf{Length} & \textbf{Time to target (s)} $\downarrow$ \\
\midrule
\textsc{Molmo}               & 1.41          & 1.41          & 75.2          & 15.4          & \phantom{0}8.22 \\
\textsc{Gaze-BeforeFirstHit} & 1.11          & \textbf{0.81} & 73.3          & \phantom{0}4.2 & \phantom{0}4.86 \\
\textsc{Gaze-SeqAnyHit}      & \textbf{0.88} & 0.95          & 80.0          & \phantom{0}9.5 & \phantom{0}7.04 \\
\textsc{Gaze-SeqLPHit}       & 1.07          & 1.03          & 77.7          & 11.0          & \phantom{0}8.29 \\
\textsc{Gaze-Shaping}        & \underline{0.99} & 0.87          & 75.5          & \phantom{0}4.0 & \phantom{0}5.13 \\
\textsc{REC-BeforeFirstHit}  & 1.49          & 0.99          & 66.8          & \phantom{0}2.0 & \phantom{0}3.98 \\
\textsc{REC-SeqAnyHit}       & 1.81          & 1.85          & 85.0          & 24.4          & 11.26 \\
\textsc{REC-Success}         & 1.29          & 1.41          & 86.0 & 18.1          & \phantom{0}8.79 \\
\textsc{REC-Shaping}         & 1.08          & \underline{0.86}          & 73.9 & \phantom{0}4.4          & \phantom{0}4.97 \\
\midrule
\textsc{Human}               & 0.00          & 0.00          & 92.5 & \phantom{0}3.4 & \phantom{0}3.87 \\
\bottomrule
\end{tabularx}
\caption{Human evaluation averaged over all datasets with $d_{\pi^s}$ from Qwen2.5-VL for comparison. Time to target refers to the average time after speech onset when the human evaluators can accurately identify the target by clicking on it. }
\label{tab:human_eval}
\end{table}

\subsection{Evaluation}
Our goal is to train VLMs to generate referring expressions that are both \textit{effective} at communicating an intended referent, and \textit{efficient} in doing so. 
We report two core metrics: rate of communicative success and reference length.
We evaluate across four referring expression datasets, using both human and neural model listeners to measure communicative success.

\noindent\textbf{Metrics} For each test dataset $\mathcal{D}^\text{test}_\text{I} = \left\{\mathcal{I}^{(i)}, b^{(i)}, \hat{x}^{(i)}\right\}_{i=1}^{D_I}$ with $D_I$ examples (each also containing a human-written reference $\hat{x}^{(i)}$), a speaker policy $\pi^\text{S}$, and listener policy $\pi^\text{REC}$, we report the \textit{rate of communicative success} 
as the proportion of examples in the test set for which the sampled referring expression $x^{(i)}$ results in the listener policy $\pi^\text{REC}$ identifying the target reference circumscribed by $b^{(i)}$ in image $\mathcal{I}^{(i)}$.
We compute the average \textit{reference length} as the mean number of words.
During development of system variants and when choosing hyperparameters, and before deploying models as speakers in interaction with human listeners, we target a metric $d_{\pi^\text{S}}$ that combines both core metrics using their  distance to the results of human-written references relative to the base model, \textsc{Molmo}. 
Metrics for human-written references are computed as:
\begin{small}
\begin{align*}
    A_{\textsc{Human}} = \dfrac{1}{D_I} \sum_{i=1}^{D_I} \mathbb{I}\left[\arg \max \pi^\text{REC}\left(\hat{x}^{(i)}, \mathcal{I}^{(i)} \right) \in b^{(i)}\right] && L_\textsc{Human} = \dfrac{1}{D_I} \sum_{i=1}^{D_I} \left|\hat{x}^{(i)}\right| \; \; .
\end{align*}
\end{small}
The normalized distance to the upper-bound performance for some policy $\pi^\text{S}$ is then
\begin{small}$$d_{\pi^\text{S}} = \sqrt{\left( \max\left(0, \frac{A_{\textsc{Human}} - A_{\pi^\text{S}}}{A_{\textsc{Human}} - A_{\textsc{Molmo}}}\right) \right)^2 + \left( \max\left(0, \frac{L_{\pi^\text{S}} - L_{\textsc{Human}}}{L_{\textsc{Molmo}} - L_{\textsc{Human}}}\right) \right)^2} \; \; .$$\end{small}
When reporting results, we use three random seeds to train three models per system and report the average on all metrics across the three trials.

\noindent\textbf{Datasets} We evaluate generated references on four different test sets $\mathcal{D}^\text{test}_\text{I}$: (a) RefCOCO Test A (1,975 examples), a subset of the RefCOCO test set whose target referents are people; (b) RefCOCO Test B (1,810 examples), a subset of the RefCOCO test set whose target referents are not people; (c) RefOI Co-Occurrence (993 examples) \citep{Ma2025-pd}, a benchmark designed to elicit referring expressions that successfully distinguish between multiple objects resembling the target referent; and (d) RefOI Single Presence (492 examples), which contains simple images that contain a single salient target referent and few to no distractors.

\noindent\textbf{Listener policies} We evaluate speaker policy performance in interaction with two listener policies $\pi^\text{REC}$: \textsc{QwenVL} and \textsc{Human}.
\textsc{QwenVL} uses the Qwen-VL-2.5-7B model \citep{bai2025qwen25vltechnicalreport} to map from an image and referring expression to a bounding box $\hat{b}$.\footnote{Although CogVLM-Grounding was found to be the best-performing REC model in \cite{Ma2025-pd}, we find it underperforms compared to Qwen-VL (see \ref{cogvlm_qwenvl}).}
We consider the proposed bounding box $\hat{b}$ to be successful if its intersection over union (IoU) with the target bounding box $b$ is greater than 0.5. 
\textsc{Human} reflects the performance of speaker models through interaction with human listeners. First, we sample 200 examples from each of the four test sets. In each trial, the image is displayed, and the referring expression is verbalized using Web Speech API's Speech Synthesis \footnote{\url{https://developer.mozilla.org/en-US/docs/Web/API/SpeechSynthesis}}. Participants are instructed to click on the region of the image they believe to be described by what they hear. If a participant clicks an image multiple times, we take their final click to be their guess at the target referent. Participants are also given the option to reveal the text of the reference after the audio has played. We collect three responses $\mathcal{P} = \{ p_1, p_2, p_3\}$ from three participants per image-reference pair; we consider the set of responses $\mathcal{P} \in b$ if the majority of guesses are in the bounding box; i.e., $\sum_{p \in \mathcal{P}} \mathbb{I}(p \in b) \geq 2$ . All three annotators for an example agreed on 77.4\% of examples. Statistics on the rate of early clicks and text reveals can be found in Appendix~\ref{additional_res} and the full details and annotator instructions are in Appendix~\ref{humanevaldetails}.

\section{Results}
\begin{figure}[b]
\label{fig:communicative_efficiency}
\begin{center}
   \includegraphics[width=\linewidth]{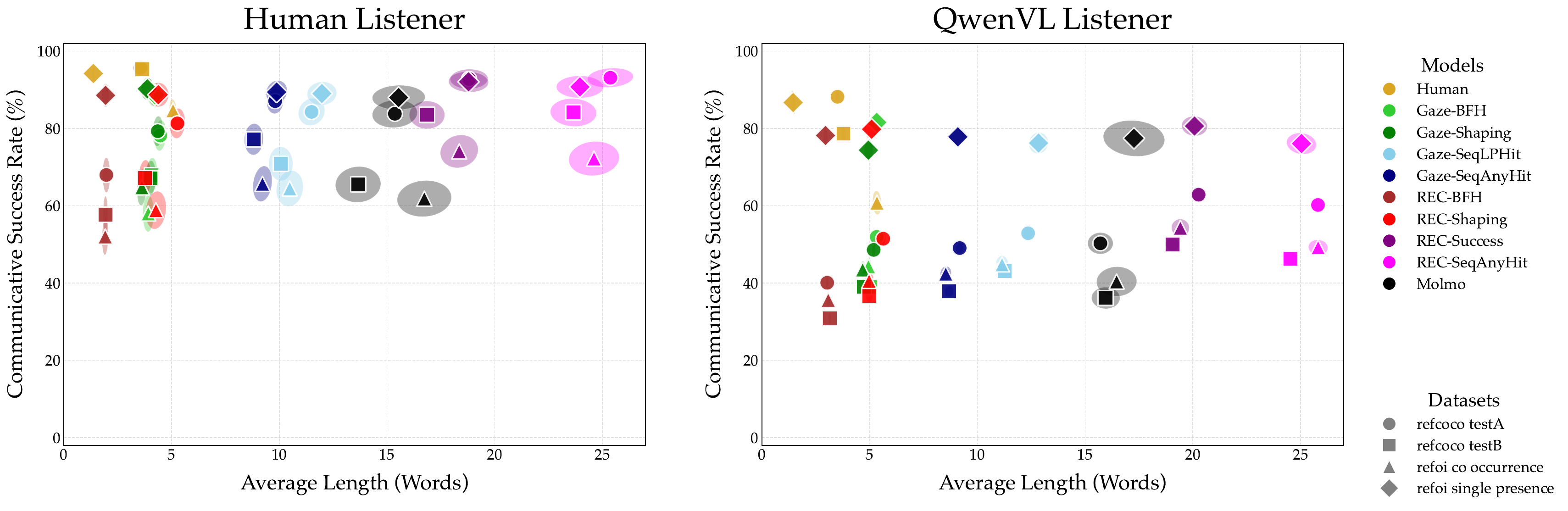}
\end{center}
\caption{Comparison of communicative success and average reference length in words with human and Qwen2.5-VL listeners with 95\% confidence ellipses.}
\end{figure}

Table~\ref{tab:human_eval} shows results for all system variations.
Models trained with our gaze-based listener Molmo-REC-Gaze significantly reduce the gap $d_{\pi^\text{S}}$ from \textsc{Molmo} to the performance of \textsc{Human} references.
Almost all gaze-based models result in references significantly shorter and referential success around or higher than \textsc{Molmo}, demonstrating that learning from listeners that estimate the process of reference comprehension yields improvements in pragmatic competence.
When paired with human listeners, these models achieve up to an 4.8\% absolute improvement in communicative success (\textsc{Gaze-SeqAnyHit}) and a reduction from 15.4 down to 4.0 words per reference (\textsc{Gaze-Shaping}).
On the other hand, while speakers trained with non-gaze-based listeners achieve even higher boosts in communicative success up to a 10.8\% absolutely improvement (\textsc{REC-Success}), this comes at a serious cost to reference efficiency, up to references with an average of 24.4 words (\textsc{REC-SeqAnyHit}).
This is also reflected in absolute time for human listeners to identify target referents; while \textsc{REC-SeqAnyHit} outperforms \textsc{Gaze-SeqAnyHit} in communicative success by 5\%, identifying target referents takes more than 4 seconds on average longer than \textsc{Gaze-SeqAnyHit}.
Although \textsc{REC-Shaping} performs competitively with the machine listener, the communicative success suffers with human listeners, indicating the presence of some behavior that is less intelligible to human listeners. Further discussion of such behaviors can be found in Appendix~\ref{sec:qualitative_error_analysis}.
Our results illustrate a tradeoff between efficiency and efficacy in reference generation (Figure~\ref{fig:communicative_efficiency}).
Overall, we find that training with listeners that estimate the process of reference comprehension, rather than only the outcome, results in speakers that generate much more humanlike references.\footnote{Full results per dataset and all model-based evaluation can be found in Appendix~\ref{additional_res}}

\label{fig:refexamples}
\begin{figure}[t]
\begin{center}
   \includegraphics[width=\linewidth]{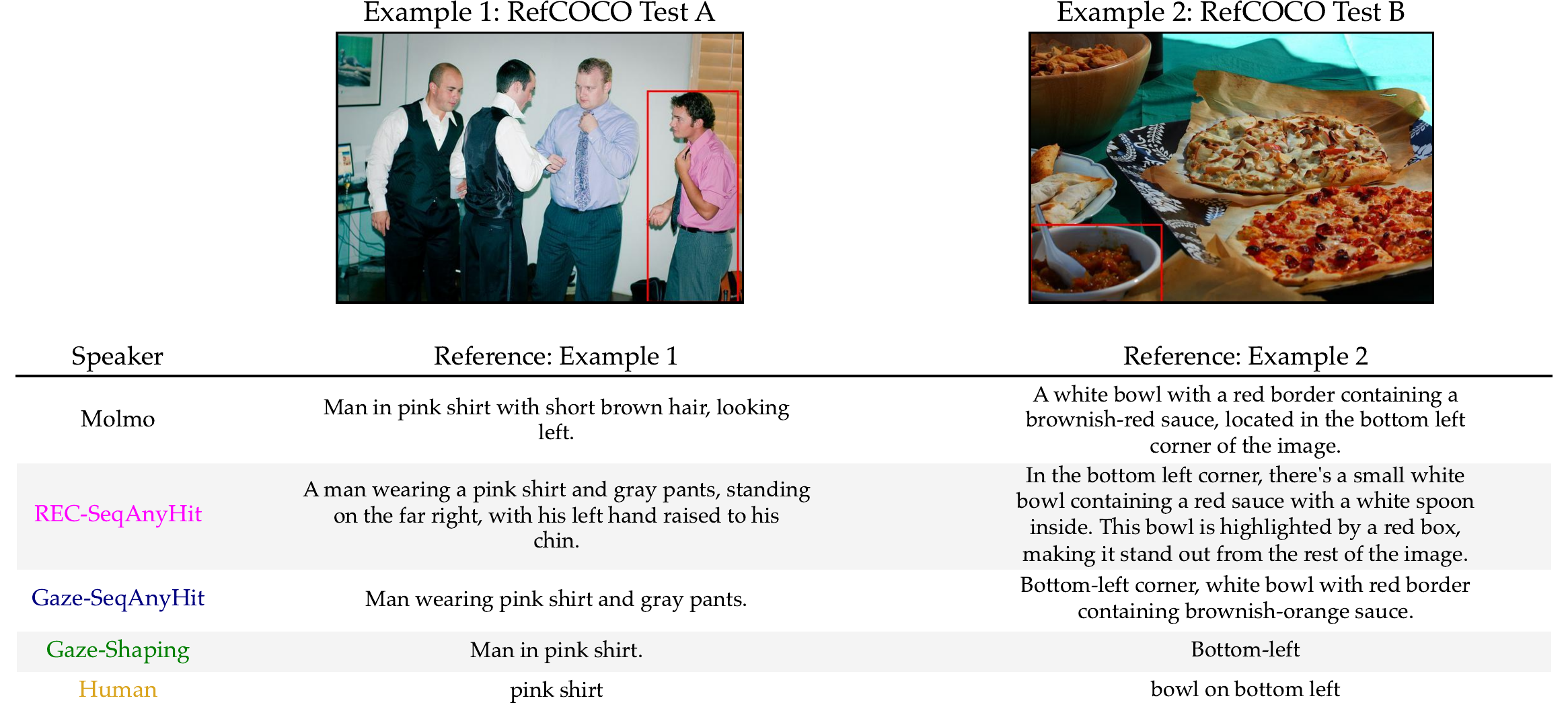}
\end{center}
\caption{Examples of generated referring expressions conditioned on an image and bounding box from several different speaker policies. See Appendix~\ref{sec:qualitative_error_analysis} for examples from all speaker models.}
\end{figure}

\begin{figure}[t]
    \centering
    \includegraphics[width=\linewidth]{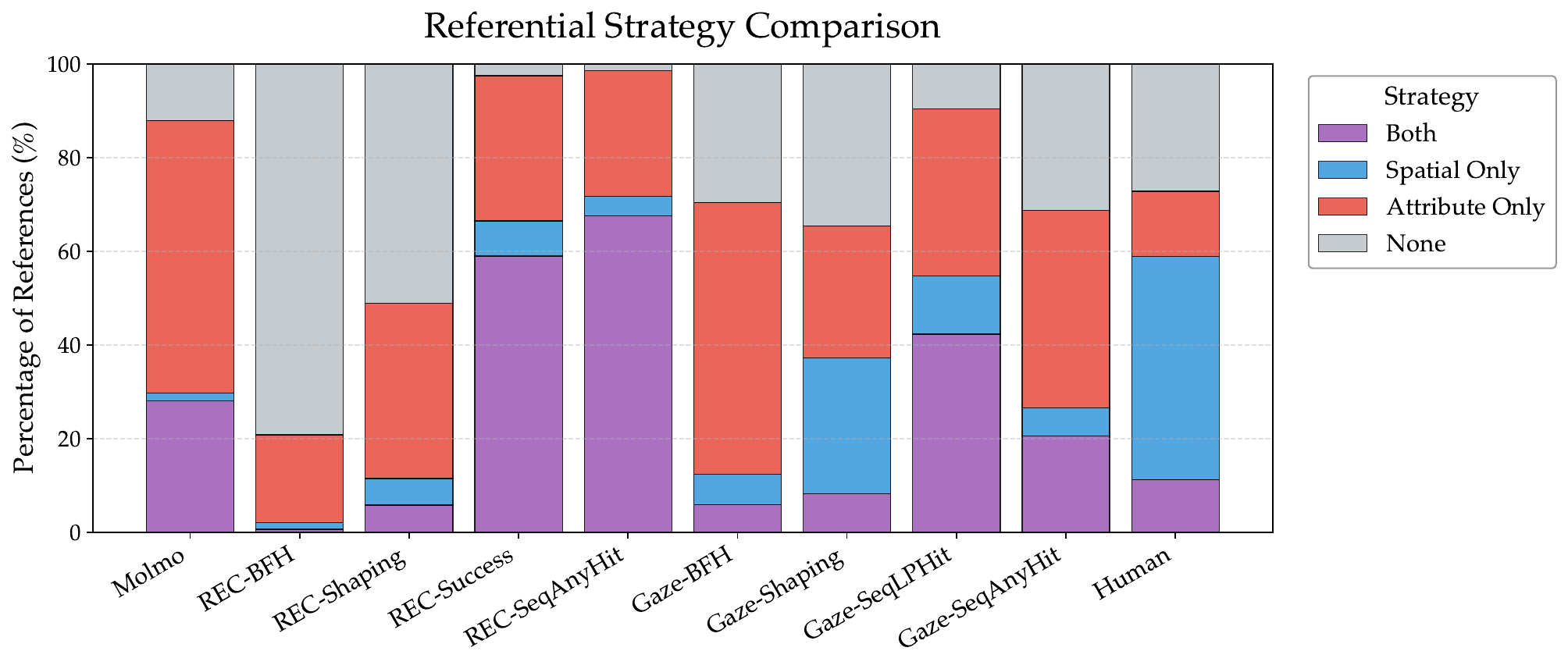}
    \caption{Comparison of referential strategies utilized by different speakers. The keywords used for this analysis can be found in Appendix~\ref{appx:refstrategy}.}
    \label{fig:strategy}
\end{figure}

Beyond main results, we analyze the shifts in referential behavior induced by gaze-based and REC-based learning. We consider the referential strategy of each speaker model using a list of keywords to tag each generated reference as utilizing spatial reference, attribute-based reference, or both (Figure \ref{fig:strategy}). We find that the \textsc{Gaze-Shaping} policy is the only machine speaker to use mostly spatial reference; its distribution of referential strategies most closely resembles the human distribution. While the accuracy and length of references from \textsc{Gaze-Shaping} and \textsc{REC-Shaping} are similar for machine listeners, human evaluation and the differences in referential strategy demonstrate that gaze-based rewards result in more humanlike references.

We also conduct an analysis of the patterns observed in \cite{Ma2025-pd}, noting what portions of references are redundant, or if the reference is ambiguous or misleading for a subset of 50 examples (Appendix~\ref{subsec:refquality}). Our analysis reveals that gaze-based optimization reduces redundancy, but also produces more ambiguous expressions. While the \textsc{Molmo} baseline and \textsc{REC-Success} policies generate an average of 13.35 and 15.01 unnecessary words per reference, respectively, \textsc{Gaze-SeqAnyHit} nearly halves this redundancy to 7.85 words, and \textsc{Gaze-Shaping} further minimizes it to just 3.26 words. 

\section{Conclusion}
In this work, we demonstrate that VLMs can be finetuned to generate more pragmatically optimal referring expressions using gaze as an implicit signal of listener comprehension. Policies that learn from gaze reduce the gap between base VLM outputs and human-like performance, reducing average length by as much as 11.4 words while maintaining or improving communicative success rates. Furthermore, the \textsc{Gaze-Shaping} policy is the only speaker policy to prioritize spatial reference, mirroring human behavior without explicit feedback. These results demonstrate that learning from signals provided from a human-like listener result in more human-like generations. Nonetheless, a tradeoff remains; speaker policies trained with REC listeners achieve the highest accuracy rates, but do so by producing verbose references. Overall, our findings establish that gaze-based insights into the process of comprehension offer a valuable implicit signal for advancing pragmatic language generation.


\section*{Acknowledgments}
This material is based upon work supported by the National Science Foundation Graduate Research Fellowship Program under Grant No. DGE 2146752. Any opinions, findings, and conclusions or recommendations expressed in this material are those of the author(s) and do not necessarily reflect the views of the National Science Foundation.
This research was
supported by an Amazon Research Award, a gift from Google, a Technical AI Safety Research award
from Coefficient Giving, and an NVIDIA Academic Grant Program award.

\section*{Ethics Statement}
This work focuses exclusively on English, potentially reinforcing linguistic patterns and cultural norms from high-resource languages; our approach may inadvertently marginalize low-resource or non-standard dialects. We acknowledge that generating human-like expressions requires ongoing oversight to prevent the reinforcement of systemic linguistic biases.

All human evaluation followed ethical research practices, with 586 participants completing the study after quality filtering. Participants were compensated at an average rate of \$19.67 USD per hour. Interaction data, including clicks and timestamps, was fully anonymized to protect participant privacy.

\bibliography{colm2026_conference}
\bibliographystyle{colm2026_conference}

\appendix

\section{Gaze prediction details and evaluation} \label{appx:gazepred}

\subsection{Point format augmentation}
For the Molmo-based architectures, we augment the standard point format to exclude alternate text and quotation marks to ensure simplicity during reward calculation. The Molmo point formatting
\begin{quote}
    $<\text{point x="50.0" y="50.0" alt="cat"}>\text{cat}<\text{/point}>$
\end{quote}
is simplified during finetuning the gaze predictor to:
\begin{quote}
    $<\text{x=50.0 y=50.0}>$
\end{quote}
This formatting change also applies to other finetuned listener models trained for ablations in following appendices.  

\subsection{Example of gaze prediction}

\begin{figure}[h]
\begin{center}
    \includegraphics[width=0.5\linewidth]{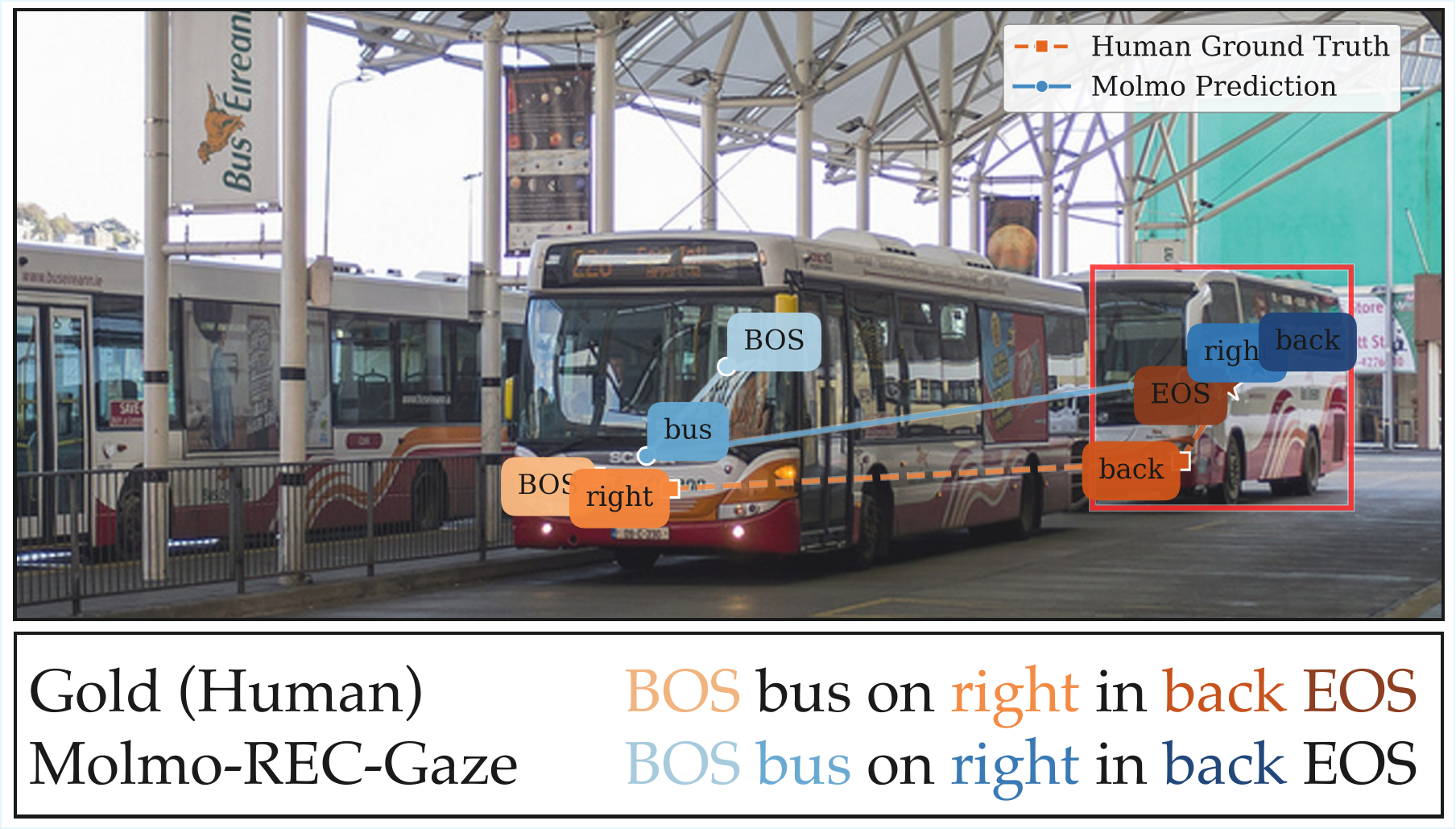}
\end{center}
\caption{Example of the iterative gaze prediction for gold and finetuned Molmo. The orange sequence represents the human scanpath and the blue is that produced by our gaze predictor model. Each point indicates the location of the gaze and the word heard at the time the fixation was recorded.}
\label{fig:gaze-est-ex}
\end{figure}

\subsection{Full gaze prediction results}
Below are the full evaluation results for our gaze predictor and ART \citep{Mondal2024-ul}. We include the distance to the gold point both averaged over all gold scanpaths and using the closest gold scanpath at each word index and total. In both cases, the VLM-based predictor is closer to the human scanpath than ART.
\begin{table}[h]
\centering
\begin{tabular}{lcc}
\toprule
\textbf{Word Index} & \textbf{Molmo Avg Dist} & \textbf{ART Avg Dist} \\ 
\midrule
0                    & 4.25                     & 4.43                  \\
1                    & 12.17                    & 9.81                  \\
2                    & 12.85                    & 12.56                 \\
3                    & 13.59                    & 14.08                 \\
4                    & 12.10                    & 11.56                 \\
5                    & 12.77                    & 13.11                 \\
6                    & 5.28                     & 12.83                 \\
7                    & 13.12                    & 8.83                  \\
8                    & 3.21                     & 10.34                 \\
\midrule
\textbf{Overall}    & \textbf{62.24}          & 72.86                \\ 
\bottomrule
\end{tabular}
\caption{Average Euclidean distance per word index and total sum of distances, averaged over all gold sequences.}
\label{tab:dist_avg_all}
\end{table}

\begin{table}[H]
\centering
\begin{tabular}{lcc}
\toprule
\textbf{Word Index} & \textbf{Molmo Avg Dist} & \textbf{ART Avg Dist} \\ 
\midrule
0                    & 2.04                     & 2.01                  \\
1                    & 10.56                    & 6.59                  \\
2                    & 8.63                     & 6.54                  \\
3                    & 9.92                     & 8.56                  \\
4                    & 9.39                     & 7.70                  \\
5                    & 11.78                    & 9.99                  \\
6                    & 2.70                     & 7.23                  \\
7                    & 7.34                     & 7.87                  \\
8                    & 3.31                     & 11.22                 \\ 
\midrule
\textbf{Overall}    & \textbf{43.21}          & 45.75                \\ 
\bottomrule
\end{tabular}
\caption{Average Euclidean distance per word index and total sum of distances, using the closest gold path (Best Match).}
\label{tab:dist_best_match}
\end{table}

\subsection{Predicting gaze with human and machine-generated references}
Since the gaze prediction model is trained only on human-generated references, we analyze differences in gaze prediction accuracy between the gold references and machine-generated references resembling those the model would see at the start of training. To do so, we sample 100 examples from each of the four evaluation datasets and generate scanpaths for these 400 examples using the gold references for each as well as references generated by Molmo-7B with no additional training. 

Across human and machine-generated references, the gaze predictor achieves higher accuracy and fewer null points on shorter references, although it is difficult to disentangle the length or speaker from the difficulty of the produced references. The difference in performance definitely influences the resulting speaker policies trained with the gaze listener. However, it is intentional that the listener used during speaker training reflects a human listener’s reference resolution process, in order to result in a speaker that generates references that are easy for a human listener to understand. The Molmo REC listener excels at reference resolution, in that it is trained to understand references that humans cannot (i.e., coordinates). Our results show that a listener trained on the kinds of expressions people use, and the processes involved in resolving such references, results in the production of more humanlike references. 
\begin{figure}[h]
\begin{center}
    \includegraphics[width=\linewidth]{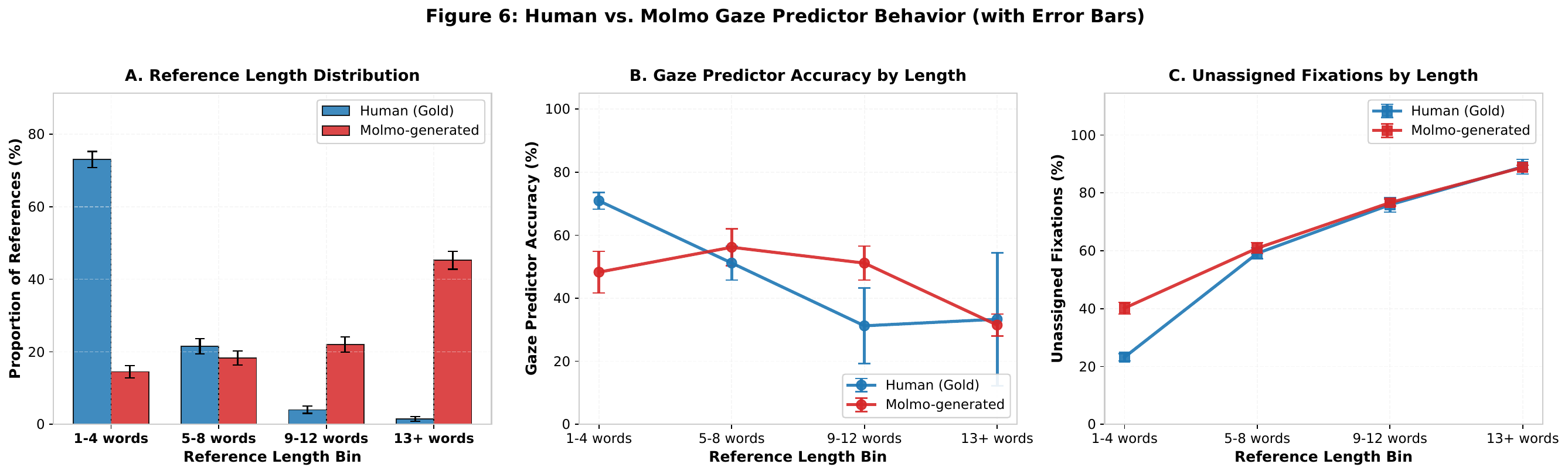}
\end{center}
\caption{(A) Comparison of length distributions of human and Molmo-generated references, (B) accuracy of the gaze predictor on human-generated (blue) and Molmo-generated references (red) over different reference lengths, and (C) comparison of the proportion of words not assigned fixations.}
\end{figure}

\section{Full REG results and evaluator choice} \label{additional_res}

\subsection{Full REG results}
In this section, we report the complete dataset-level evaluation for all speaker variants. Table~\ref{tab:appendix_eval_grouped} details the REF performance metrics: accuracy, average utterance length, average time for a human listener to resolve the expression, and $d_{\pi^s}$. Additionally, Table~\ref{tab:appendix_human_behavioral} presents behavioral metrics--- resolution time, reveal rate, and early click rate. Finally, we present the averaged and per-dataset results using the QwenVL listener in Tables \ref{tab:appendix_qwen_eval} and \ref{tab:avg_eval_qwen}, respectively.
\begin{table}[H]
\centering
\small
\begin{tabular}{lcccc}
\toprule
\textbf{Speaker} & \textbf{Acc.} $\uparrow$ & \textbf{Length} & \textbf{Time (s)} $\downarrow$ & \textbf{$d_{\pi^s}$} $\downarrow$ \\
\midrule
\multicolumn{5}{c}{\textit{RefCOCO testA}} \\
\midrule
\textsc{Human (Gold)}          & 95.67 & \phantom{0}3.59 & \phantom{0}3.97 & 0.00 \\
\textsc{Molmo}                 & 84.17 & 15.15           & \phantom{0}7.99 & 1.41 \\
\textsc{REC-BeforeFirstHit}    & 68.50 & \phantom{0}2.01 & \phantom{0}4.23 & 2.37 \\
\textsc{REC-SeqAnyHit}         & 93.17 & 25.27           & 11.40           & 1.89 \\
\textsc{REC-Success}          & 93.50 & 18.58           & \phantom{0}9.21 & 1.31 \\
\textsc{REC-Shaping}          & 81.17 & \phantom{0}5.27 & \phantom{0}5.30 & 1.27 \\
\textsc{Gaze-SeqAnyHit}        & 87.83 & \phantom{0}9.87 & \phantom{0}6.35 & 0.87 \\
\textsc{Gaze-SeqLPHit}         & 85.00 & 11.53           & \phantom{0}8.94 & 1.15 \\
\textsc{Gaze-Shaping}          & 79.83 & \phantom{0}4.36 & \phantom{0}4.52 & 1.38 \\
\textsc{Gaze-BeforeFirstHit}  & 79.00 & \phantom{0}4.48 & \phantom{0}5.12 & 1.45 \\
\midrule
\multicolumn{5}{c}{\textit{RefCOCO testB}} \\
\midrule
\textsc{Human (Gold)}          & 96.17 & \phantom{0}3.73 & \phantom{0}4.45 & 0.00 \\
\textsc{Molmo}                 & 65.50 & 13.84           & \phantom{0}8.19 & 1.41 \\
\textsc{REC-BeforeFirstHit}    & 58.17 & \phantom{0}1.93 & \phantom{0}3.99 & 1.25 \\
\textsc{REC-SeqAnyHit}         & 83.83 & 23.56           & 11.98           & 2.00 \\
\textsc{REC-Success}          & 83.67 & 16.90           & \phantom{0}8.76 & 1.37 \\
\textsc{REC-Shaping}          & 67.00 & \phantom{0}3.78 & \phantom{0}5.20 & 0.95 \\
\textsc{Gaze-SeqAnyHit}        & 77.33 & \phantom{0}8.84 & \phantom{0}7.13 & 0.80 \\
\textsc{Gaze-SeqLPHit}         & 71.33 & 10.08           & \phantom{0}8.04 & 1.02 \\
\textsc{Gaze-Shaping}          & 67.33 & \phantom{0}4.06 & \phantom{0}5.44 & 0.94 \\
\textsc{Gaze-BeforeFirstHit}  & 66.83 & \phantom{0}4.04 & \phantom{0}5.17 & 0.96 \\
\midrule
\multicolumn{5}{c}{\textit{RefOI Co-occurrence}} \\
\midrule
\textsc{Human (Gold)}          & 84.83 & \phantom{0}5.02 & \phantom{0}4.07 & 0.00 \\
\textsc{Molmo}                 & 62.83 & 17.00           & \phantom{0}9.45 & 1.41 \\
\textsc{REC-BeforeFirstHit}    & 52.00 & \phantom{0}1.98 & \phantom{0}4.08 & 1.51 \\
\textsc{REC-SeqAnyHit}         & 72.50 & 24.70           & 11.50           & 1.74 \\
\textsc{REC-Success}          & 74.67 & 18.36           & \phantom{0}8.66 & 1.20 \\
\textsc{REC-Shaping}          & 59.00 & \phantom{0}4.21 & \phantom{0}5.51 & 1.18 \\
\textsc{Gaze-SeqAnyHit}        & 65.83 & \phantom{0}9.24 & \phantom{0}7.24 & 0.93 \\
\textsc{Gaze-SeqLPHit}         & 65.17 & 10.48           & \phantom{0}7.78 & 1.00 \\
\textsc{Gaze-Shaping}          & 64.33 & \phantom{0}3.63 & \phantom{0}5.35 & 0.94 \\
\textsc{Gaze-BeforeFirstHit}  & 58.33 & \phantom{0}3.86 & \phantom{0}5.28 & 1.21 \\
\midrule
\multicolumn{5}{c}{\textit{RefOI Single Presence}} \\
\midrule
\textsc{Human (Gold)}          & 93.50 & \phantom{0}1.39 & \phantom{0}3.28 & 0.00 \\
\textsc{Molmo}                 & 88.33 & 15.57           & \phantom{0}7.57 & 1.41 \\
\textsc{REC-BeforeFirstHit}    & 88.33 & \phantom{0}1.98 & \phantom{0}3.73 & 1.00 \\
\textsc{REC-SeqAnyHit}         & 90.67 & 24.00           & 10.26           & 1.69 \\
\textsc{REC-Success}          & 92.00 & 18.68           & \phantom{0}8.51 & 1.25 \\
\textsc{REC-Shaping}          & 88.33 & \phantom{0}4.45 & \phantom{0}4.14 & 1.02 \\
\textsc{Gaze-SeqAnyHit}        & 89.00 & \phantom{0}9.94 & \phantom{0}6.67 & 1.06 \\
\textsc{Gaze-SeqLPHit}         & 89.33 & 12.07           & \phantom{0}8.24 & 1.10 \\
\textsc{Gaze-Shaping}          & 90.33 & \phantom{0}3.89 & \phantom{0}4.44 & 0.64 \\
\textsc{Gaze-BeforeFirstHit}  & 88.83 & \phantom{0}4.29 & \phantom{0}4.12 & 0.93 \\
\bottomrule
\end{tabular}
\caption{Full human evaluation results by dataset. }
\label{tab:appendix_eval_grouped}
\end{table}

\clearpage
\subsection{Training curves}
\begin{figure}[H]
\begin{center}
    \includegraphics[width=0.9\linewidth]{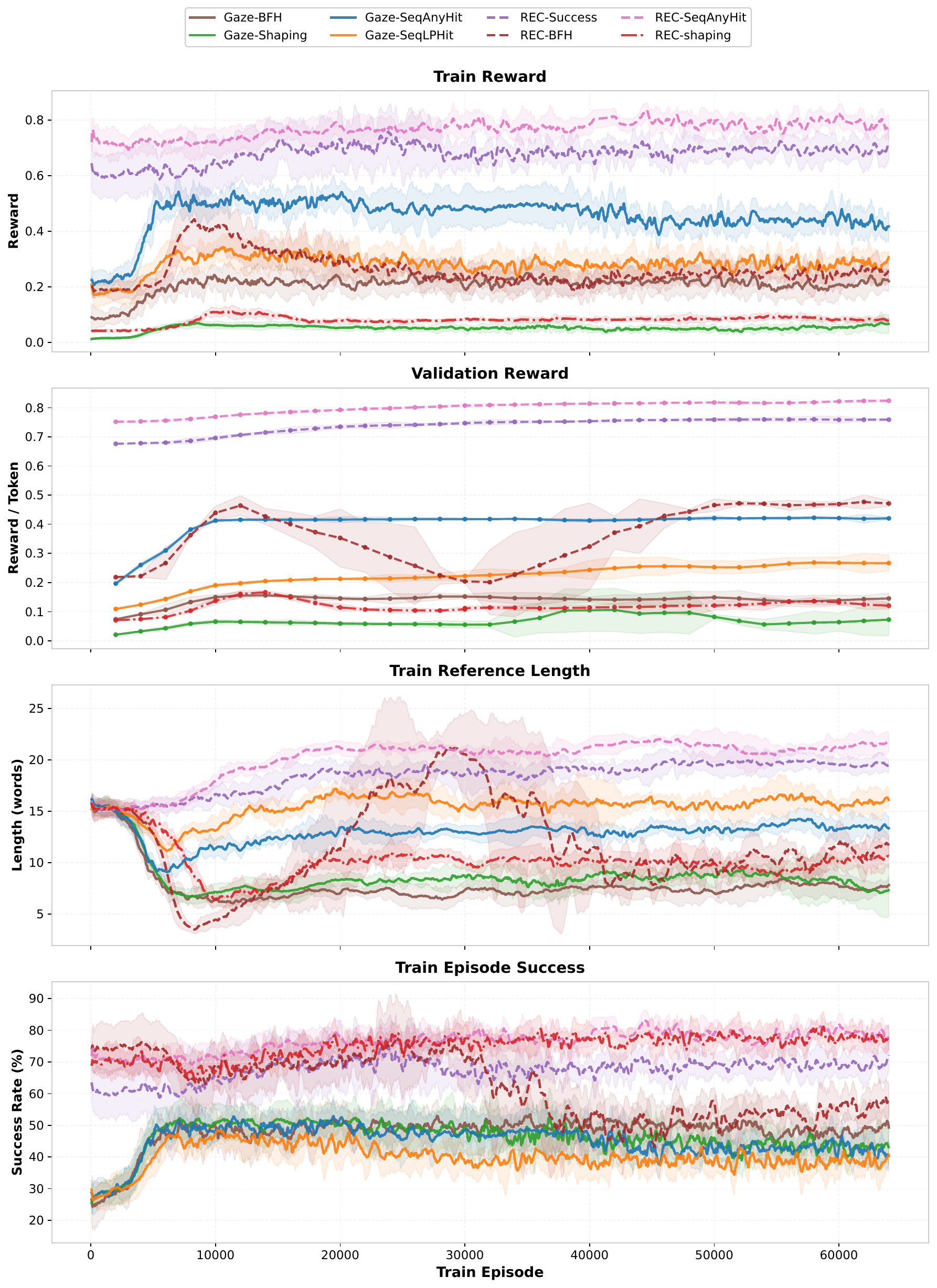}
\end{center}
\label{appx:curves}
\end{figure}

Fig~\ref{appx:curves} shows the train and validation reward, success rate of the reference (i.e., whether any point in the set of produced point was within the target region), and generated reference length in tokens. The shaded area represents the range of values over three random seeds. The rewards and success rates for gaze-based rewards tend to be lower than their REC counterparts. The \textsc{REC-BeforeFirstHit} training is noticeably the most unstable relative to the other reward configurations. 
\clearpage

\begin{table}[H]
\centering
\small
\begin{tabular}{lcccc}
\toprule
\textbf{Speaker} & \textbf{Time (s)} $\downarrow$ & \textbf{Reveal Rate} $\downarrow$ & \textbf{Early Click Rate} $\downarrow$ \\
\midrule
\multicolumn{4}{c}{\textit{Overall}} \\
\midrule
\textsc{Human (Gold)}          & \phantom{0}3.87 & 0.03 & 0.05 \\
\textsc{Molmo}                 & \phantom{0}8.22 & 0.05 & 0.27 \\
\textsc{REC-BeforeFirstHit}    & \phantom{0}3.98 & 0.04 & 0.04 \\
\textsc{REC-SeqAnyHit}         & 11.26           & 0.05 & 0.45 \\
\textsc{REC-Success}          & \phantom{0}8.79 & 0.03 & 0.41 \\
\textsc{REC-Shaping}          & \phantom{0}5.04 & 0.09 & 0.13 \\
\textsc{Gaze-SeqAnyHit}        & \phantom{0}7.04 & 0.03 & 0.21 \\
\textsc{Gaze-SeqLPHit}         & \phantom{0}8.29 & 0.04 & 0.32 \\
\textsc{Gaze-Shaping}          & \phantom{0}5.13 & 0.04 & 0.06 \\
\textsc{Gaze-BeforeFirstHit}  & \phantom{0}4.86 & 0.04 & 0.06 \\
\midrule
\multicolumn{4}{c}{\textit{RefCOCO testA}} \\
\midrule
\textsc{Human (Gold)}          & \phantom{0}3.97 & 0.03 & 0.04 \\
\textsc{Molmo}                 & \phantom{0}7.99 & 0.05 & 0.24 \\
\textsc{REC-BeforeFirstHit}    & \phantom{0}4.23 & 0.06 & 0.02 \\
\textsc{REC-SeqAnyHit}         & 11.40           & 0.05 & 0.36 \\
\textsc{REC-Success}          & \phantom{0}9.21 & 0.04 & 0.38 \\
\textsc{REC-Shaping}          & \phantom{0}5.30 & 0.07 & 0.13 \\
\textsc{Gaze-SeqAnyHit}        & \phantom{0}6.35 & 0.03 & 0.27 \\
\textsc{Gaze-SeqLPHit}         & \phantom{0}8.94 & 0.05 & 0.35 \\
\textsc{Gaze-Shaping}          & \phantom{0}4.52 & 0.05 & 0.05 \\
\textsc{Gaze-BeforeFirstHit}  & \phantom{0}5.12 & 0.04 & 0.06 \\
\midrule
\multicolumn{4}{c}{\textit{RefCOCO testB}} \\
\midrule
\textsc{Human (Gold)}          & \phantom{0}4.45 & 0.05 & 0.04 \\
\textsc{Molmo}                 & \phantom{0}8.19 & 0.05 & 0.17 \\
\textsc{REC-BeforeFirstHit}    & \phantom{0}3.99 & 0.05 & 0.04 \\
\textsc{REC-SeqAnyHit}         & 11.98           & 0.06 & 0.39 \\
\textsc{REC-Success}          & \phantom{0}8.76 & 0.05 & 0.31 \\
\textsc{REC-Shaping}          & \phantom{0}5.20 & 0.10 & 0.10 \\
\textsc{Gaze-SeqAnyHit}        & \phantom{0}7.13 & 0.05 & 0.14 \\
\textsc{Gaze-SeqLPHit}         & \phantom{0}8.04 & 0.05 & 0.28 \\
\textsc{Gaze-Shaping}          & \phantom{0}5.44 & 0.06 & 0.06 \\
\textsc{Gaze-BeforeFirstHit}  & \phantom{0}5.17 & 0.05 & 0.04 \\
\midrule
\multicolumn{4}{c}{\textit{RefOI Co-occurrence}} \\
\midrule
\textsc{Human (Gold)}          & \phantom{0}4.07 & 0.03 & 0.07 \\
\textsc{Molmo}                 & \phantom{0}9.45 & 0.04 & 0.29 \\
\textsc{REC-BeforeFirstHit}    & \phantom{0}4.08 & 0.03 & 0.06 \\
\textsc{REC-SeqAnyHit}         & 11.50           & 0.06 & 0.42 \\
\textsc{REC-Success}          & \phantom{0}8.66 & 0.03 & 0.42 \\
\textsc{REC-Shaping}          & \phantom{0}5.51 & 0.11 & 0.11 \\
\textsc{Gaze-SeqAnyHit}        & \phantom{0}7.24 & 0.04 & 0.15 \\
\textsc{Gaze-SeqLPHit}         & \phantom{0}7.78 & 0.06 & 0.26 \\
\textsc{Gaze-Shaping}          & \phantom{0}5.35 & 0.04 & 0.03 \\
\textsc{Gaze-BeforeFirstHit}  & \phantom{0}5.28 & 0.04 & 0.05 \\
\midrule
\multicolumn{4}{c}{\textit{RefOI Single Presence}} \\
\midrule
\textsc{Human (Gold)}          & \phantom{0}3.28 & 0.03 & 0.05 \\
\textsc{Molmo}                 & \phantom{0}7.57 & 0.05 & 0.39 \\
\textsc{REC-BeforeFirstHit}    & \phantom{0}3.73 & 0.03 & 0.06 \\
\textsc{REC-SeqAnyHit}         & 10.26           & 0.05 & 0.63 \\
\textsc{REC-Success}          & \phantom{0}8.51 & 0.02 & 0.53 \\
\textsc{REC-Shaping}          & \phantom{0}4.14 & 0.07 & 0.17 \\
\textsc{Gaze-SeqAnyHit}        & \phantom{0}6.67 & 0.03 & 0.28 \\
\textsc{Gaze-SeqLPHit}         & \phantom{0}8.24 & 0.03 & 0.39 \\
\textsc{Gaze-Shaping}          & \phantom{0}4.44 & 0.03 & 0.10 \\
\textsc{Gaze-BeforeFirstHit}  & \phantom{0}4.12 & 0.02 & 0.11 \\
\bottomrule
\end{tabular}
\caption{Behavioral efficiency metrics grouped by dataset, including overall averages.}
\label{tab:appendix_human_behavioral}
\end{table}

\begin{table}[H]
\centering
\small
\begin{tabular}{lcccc}
\toprule
\textbf{Speaker} & \textbf{Acc.} $\uparrow$ & \textbf{Length} & \textbf{$d_{\pi^s}$} $\downarrow$ \\
\midrule
\multicolumn{4}{c}{\textit{RefCOCO testA}} \\
\midrule
\textsc{Human (Gold)}          & 94.60 & \phantom{0}3.51 & 0.00 \\
\textsc{Molmo}                 & 48.20 & 15.71           & 1.41 \\
\textsc{REC-BeforeFirstHit}    & 40.07 & \phantom{0}3.03 & 1.18 \\
\textsc{REC-SeqAnyHit}         & 60.20 & 25.81           & 1.97 \\
\textsc{REC-Success}          & 62.85 & 20.27           & 1.53 \\
\textsc{REC-Shaping}          & 51.46 & \phantom{0}5.63 & 0.95 \\
\textsc{Gaze-SeqAnyHit}        & 49.05 & \phantom{0}9.18 & 1.09 \\
\textsc{Gaze-SeqLPHit}         & 52.88 & 12.35           & 1.16 \\
\textsc{Gaze-Shaping}          & 48.57 & \phantom{0}5.19 & 1.00 \\
\textsc{Gaze-BeforeFirstHit}  & 51.93 & \phantom{0}5.33 & 0.93 \\
\midrule
\multicolumn{4}{c}{\textit{RefCOCO testB}} \\
\midrule
\textsc{Human (Gold)}          & 88.00 & \phantom{0}3.78 & 0.00 \\
\textsc{Molmo}                 & 33.50 & 15.96           & 1.41 \\
\textsc{REC-BeforeFirstHit}    & 30.87 & \phantom{0}3.14 & 1.05 \\
\textsc{REC-SeqAnyHit}         & 46.32 & 24.52           & 1.87 \\
\textsc{REC-Success}          & 50.02 & 19.06           & 1.44 \\
\textsc{REC-Shaping}          & 36.67 & \phantom{0}4.98 & 0.95 \\
\textsc{Gaze-SeqAnyHit}        & 37.82 & \phantom{0}8.69 & 1.01 \\
\textsc{Gaze-SeqLPHit}         & 43.09 & 11.27           & 1.03 \\
\textsc{Gaze-Shaping}          & 39.01 & \phantom{0}4.73 & 0.90 \\
\textsc{Gaze-BeforeFirstHit}  & 38.99 & \phantom{0}5.01 & 0.90 \\
\midrule
\multicolumn{4}{c}{\textit{RefOI Co-occurrence}} \\
\midrule
\textsc{Human (Gold)}          & 92.90 & \phantom{0}5.34 & 0.00 \\
\textsc{Molmo}                 & 33.20 & 16.46           & 1.41 \\
\textsc{REC-BeforeFirstHit}    & 35.62 & \phantom{0}3.08 & 0.98 \\
\textsc{REC-SeqAnyHit}         & 49.22 & 25.81           & 1.98 \\
\textsc{REC-Success}          & 54.28 & 19.42           & 1.42 \\
\textsc{REC-Shaping}          & 40.55 & \phantom{0}4.97 & 0.88 \\
\textsc{Gaze-SeqAnyHit}        & 42.40 & \phantom{0}8.53 & 0.89 \\
\textsc{Gaze-SeqLPHit}         & 44.81 & 11.14           & 0.96 \\
\textsc{Gaze-Shaping}          & 43.47 & \phantom{0}4.67 & 0.83 \\
\textsc{Gaze-BeforeFirstHit}  & 44.31 & \phantom{0}4.94 & 0.81 \\
\midrule
\multicolumn{4}{c}{\textit{RefOI Single Presence}} \\
\midrule
\textsc{Human (Gold)}          & 80.40 & \phantom{0}1.45 & 0.00 \\
\textsc{Molmo}                 & 67.70 & 17.27           & 1.41 \\
\textsc{REC-BeforeFirstHit}    & 78.18 & \phantom{0}2.95 & 0.20 \\
\textsc{REC-SeqAnyHit}         & 76.08 & 25.04           & 1.53 \\
\textsc{REC-Success}          & 80.62 & 20.07           & 1.18 \\
\textsc{REC-Shaping}          & 79.81 & \phantom{0}5.09 & 0.23 \\
\textsc{Gaze-SeqAnyHit}        & 77.85 & \phantom{0}9.09 & 0.52 \\
\textsc{Gaze-SeqLPHit}         & 76.22 & 12.84           & 0.79 \\
\textsc{Gaze-Shaping}          & 74.39 & \phantom{0}4.94 & 0.52 \\
\textsc{Gaze-BeforeFirstHit}  & 81.57 & \phantom{0}5.33 & 0.26 \\
\bottomrule
\end{tabular}
\caption{QwenVL-based evaluation results by dataset.}
\label{tab:appendix_qwen_eval}
\end{table}

\begin{table}[H]
\centering
\small
\begin{tabular}{lccc}
\toprule
\textbf{Speaker} & \textbf{Acc.} $\uparrow$ & \textbf{Length} & \textbf{$d_{\pi^s}$} $\downarrow$ \\
\midrule
\multicolumn{4}{c}{\textit{Overall (Averaged)}} \\
\midrule
\textsc{Molmo}                 & 45.65           & 16.35          & 1.41 \\
\textsc{REC-BeforeFirstHit}     & 46.18           & \phantom{0}3.05 & 0.99 \\
\textsc{REC-SeqAnyHit}         & 57.96           & 25.29          & 1.84 \\
\textsc{REC-Success}           & 61.94           & 19.70          & 1.41 \\
\textsc{REC-Shaping}           & 52.12           & \phantom{0}5.17 & 0.86 \\
\textsc{Gaze-SeqAnyHit}        & 51.78           & \phantom{0}8.87 & 0.95 \\
\textsc{Gaze-SeqLPHit}         & 54.25           & 11.90          & 1.03 \\
\textsc{Gaze-Shaping}          & 51.36           & \phantom{0}4.88 & 0.87 \\
\textsc{Gaze-BeforeFirstHit}   & 54.20           & \phantom{0}5.15 & 0.81 \\
\textsc{Human (Gold)}          & 88.98           & \phantom{0}3.52 & 0.00 \\
\bottomrule
\end{tabular}
\caption{QwenVL-based evaluation results averaged across all datasets.}
\label{tab:avg_eval_qwen}
\end{table}

\subsection{Comparison of machine listener models} \label{cogvlm_qwenvl}
\begin{figure}[H]
\begin{center}
    \includegraphics[width=\linewidth]{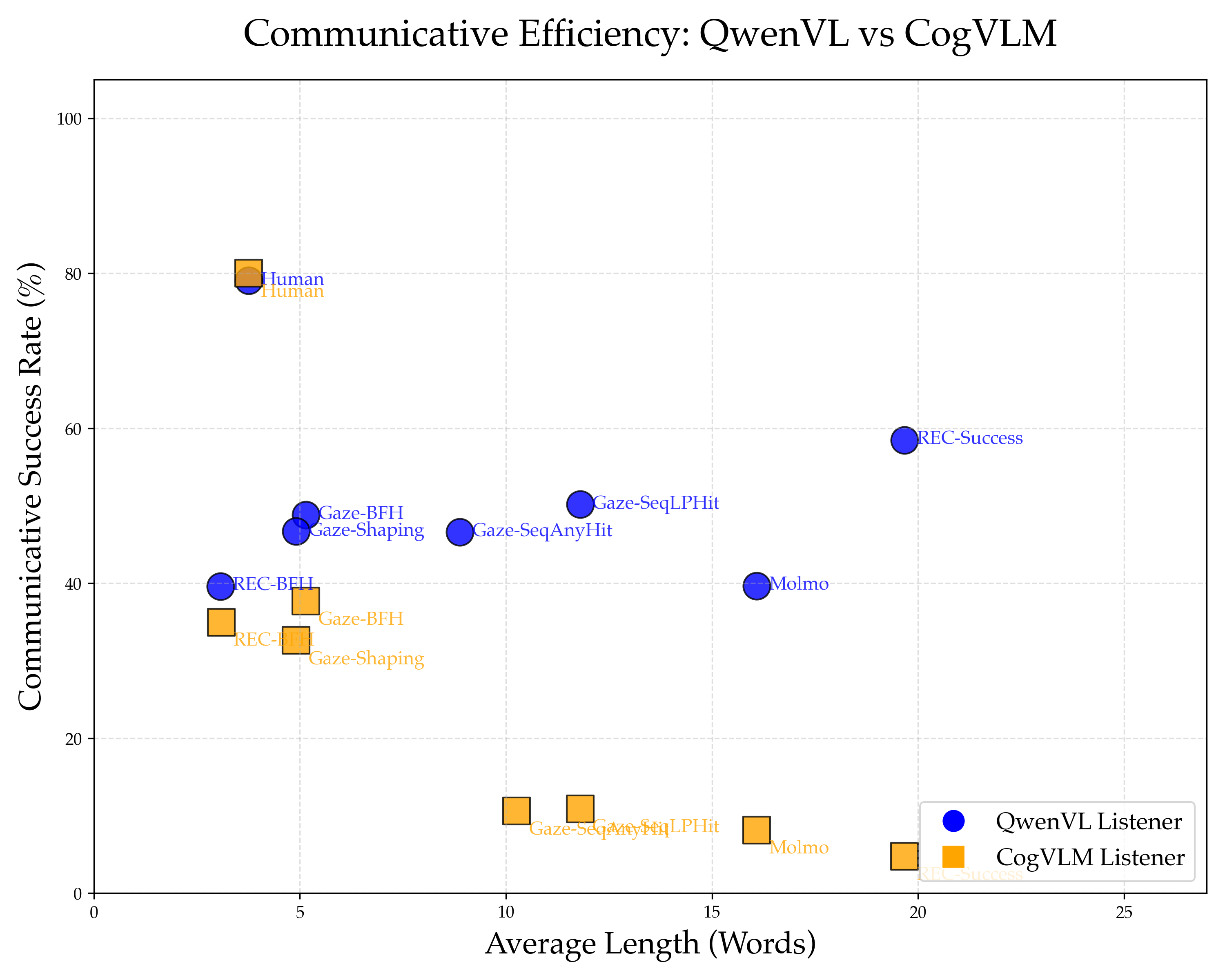}
\end{center}
\caption{Comparison of REC results with QwenVL and CogVLM-Grounding.}
\label{fig:listenercomparison}
\end{figure}

We choose to use QwenVL for our main evaluation despite \citet{Ma2025-pd} finding CogVLM-Grounding to be the best model for REC. As shown in \ref{fig:listenercomparison}, the reference resolution rate of CogVLM-Grounding becomes extremely low as references get longer, whereas QwenVL maintains higher success rates across reference lengths. We hypothesize that this difference reflects a training bias in CogVLM-Grounding toward concise, human-written references.

\section{Speaker model ablations, baselines, and comparisons}

\subsection{KL penalty ablation}
\begin{figure}[H]
\begin{center}
    \includegraphics[width=\linewidth]{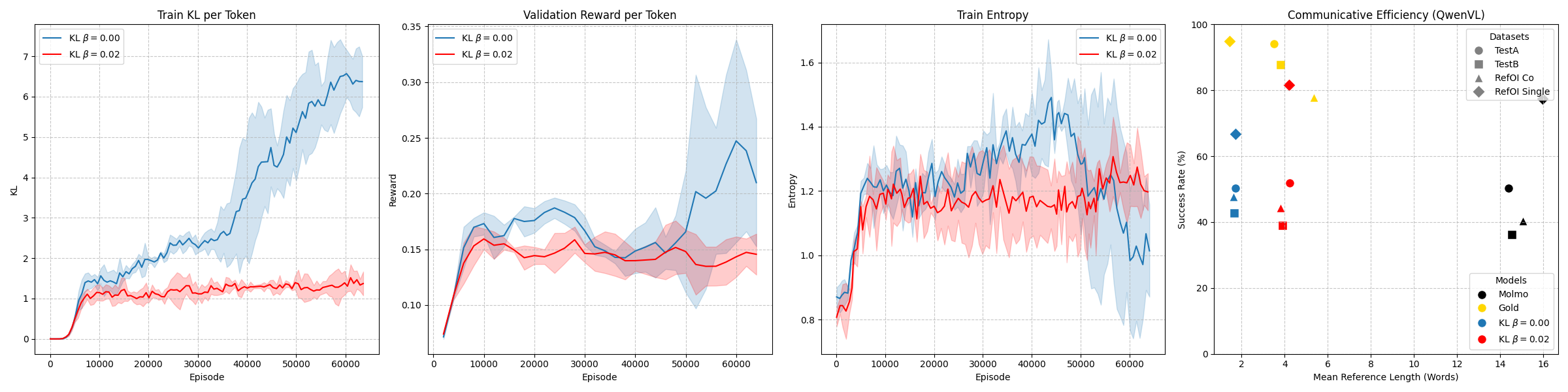}
\end{center}
\caption{Comparison of \textsc{Gaze-BFH} training performance with and without the KL penalty. Red shows training with a KL penalty ($\beta=0.02$) and blue shows the ablated training run. For comparison, we include Molmo and human speakers in the rightmost chart.}
\label{fig:kl}
\end{figure}

We demonstrate the utility of the KL penalty on the highest performing speaker model with the QwenVL listener (\textsc{Gaze-BFH}) by ablating it. Although training without the KL penalty yields higher validation rewards late in training, it causes entropy collapse and degrades overall evaluation performance (Figure~\ref{fig:kl}). With reference length collapsing to just 1--2 tokens on average, these results confirm that the KL penalty is necessary to mitigate reward hacking. As a result, we apply the KL penalty to the training of all speaker models.

\subsection{REC-Success with length penalty}
While the goal of this work is not to demonstrate that learning from gaze results in the most efficient and successful references, we compare to \textsc{REC-Success} incorporating a length penalty to control for verbosity. Because policy optimization can favor longer sequences to maximize reward, penalizing length allows for a fair comparison at closer sequence lengths. We implement the length penalty as a negative reward, $\lambda=0.05$. The penalty is applied after batch normalization to each generated token with the exception of the end-of-sequence token. The training objective becomes:
\begin{small}
\begin{multline}
\nabla_\theta \mathcal{J}(\theta) = \mathbb{E}_{(\mathcal{I}, b, x, t)} \Big( r_t - \underbrace{\beta \text{KL}\left(\pi^\text{S}_{\theta}, \mathcal{I}, b, x, t \right)}_{\text{KL penalty}} - \underbrace{\lambda \mathbb{I}(x_t \neq \text{\textsc{eos}})}_{\text{Length penalty}} \Big) \\
\times \underbrace{\nabla_\theta \left( \log \pi^\text{S}_{\theta}\left(x_t\mid\mathcal{I}, b, x_{<t}\right) \right)}_\text{Policy gradient} + \underbrace{\alpha \nabla_\theta H \left(\pi^\text{S}_{\theta}(x_t\mid\mathcal{I}, b, x_{<t})\right)}_{\text{Entropy bonus}} \; ,
\end{multline}
\end{small}
\noindent where $\mathbb{I}(\cdot)$ is the indicator function.

\begin{table}[H]
\centering
\small
\begin{tabular}{lccc}
\toprule
\textbf{Speaker} & \textbf{Acc.} $\uparrow$ & \textbf{Length} & \textbf{$d_{\pi^s}$} $\downarrow$ \\
\midrule
\multicolumn{4}{c}{\textit{Overall (Averaged)}} \\
\midrule
\textsc{Molmo}                  & 45.65 & 16.35           & 1.41 \\
\textsc{REC-BeforeFirstHit}     & 46.18 & \phantom{0}3.05 & 0.99 \\
\textsc{REC-SeqAnyHit}         & 57.96 & 25.29           & 1.84 \\
\textsc{REC-Success}           & 61.94 & 19.70           & 1.41 \\
\textsc{REC-Shaping}           & 52.12 & \phantom{0}5.17 & 0.86 \\
\textsc{Gaze-SeqAnyHit}        & 51.78 & \phantom{0}8.87 & 0.95 \\
\textsc{Gaze-SeqLPHit}         & 54.25 & 11.90           & 1.03 \\
\textsc{Gaze-Shaping}          & 51.36 & \phantom{0}4.88 & 0.87 \\
\textsc{Gaze-BeforeFirstHit}   & 54.20 & \phantom{0}5.15 & 0.81 \\
\textsc{Human (Gold)}           & 88.97           & \phantom{0}3.52 & 0.00 \\
\midrule
\multicolumn{4}{c}{\textit{Length Penalty ($\lambda=0.05$)}} \\
\midrule
\textsc{REC-Success (LP $\lambda=0.05$)} & 61.08 & \phantom{0}9.50 & 0.83 \\
\bottomrule
\end{tabular}
\caption{QwenVL-based evaluation comparing standard models against the length-penalized \textsc{REC-Success} variant.}
\label{tab:length_penalty_ablation}
\end{table}

The speaker trained with \textsc{REC-Succes} and a length penalty outperforms almost all other configurations, but falls just short of \textsc{Gaze-BeforeFirstHit}. Explicit length penalization successfully reduces verbosity, but gaze-based feedback can also encourage concise, pragmatically optimal expressions without relying on manually tuned penalty weights.

\subsection{Listener variant vaselines}

\subsubsection{Final gaze point listener}
To ensure the brevity of gaze-trained speakers is driven by the gaze signal and reward design rather than training data bias, we evaluate a speaker trained with a variant of $\pi^G$ that predicts only the final fixation point. We then integrate this model with the \textsc{REC-Success} training configuration. The resulting speaker model results in length and accuracy similar to that of \textsc{REC-Success} and even slightly longer, demonstrating that the gaze listener's training distribution is not the cause of the shorter references.

\begin{table}[H]
\centering
\small
\begin{tabular}{lccc}
\toprule
\textbf{Speaker} & \textbf{Acc.} $\uparrow$ & \textbf{Length} & \textbf{$d_{\pi^s}$} $\downarrow$ \\
\midrule
\multicolumn{4}{c}{\textit{Overall (Averaged)}} \\
\midrule
\textsc{Molmo}                 & 45.65           & 16.35          & 1.41 \\
\textsc{REC-BeforeFirstHit}     & 46.18           & \phantom{0}3.05 & 0.99 \\
\textsc{REC-SeqAnyHit}         & 57.96           & 25.29          & 1.84 \\
\textsc{REC-Success}           & 61.94           & 19.70          & 1.41 \\
\textsc{REC-Shaping}           & 52.12           & \phantom{0}5.17 & 0.86 \\
\textsc{Gaze-SeqAnyHit}        & 51.78           & \phantom{0}8.87 & 0.95 \\
\textsc{Gaze-SeqLPHit}         & 54.25           & 11.90          & 1.03 \\
\textsc{Gaze-Shaping}          & 51.36           & \phantom{0}4.88 & 0.87 \\
\textsc{Gaze-BeforeFirstHit}   & 54.20           & \phantom{0}5.15 & 0.81 \\
\textsc{Human (Gold)}          & 88.98           & \phantom{0}3.52 & 0.00 \\
\midrule
\multicolumn{4}{c}{\textit{Baseline: Final Gaze Point Listener}} \\
\midrule
\textsc{Gaze-Success (Final Gaze Point Listener)} & 61.17 & 21.47 & 1.56 \\
\bottomrule
\end{tabular}
\caption{Evaluation comparing speakers trained on full incremental gaze versus models predicting only the final gaze point.}
\label{tab:final_gaze_point_baseline}
\end{table}

\subsubsection{RefCOCO finetuned listener}
To ensure the gaze listener does not outperform the VLM listener solely through indirect exposure to RefCOCO grounding data, we finetune Molmo-7B on the RefCOCO subset within the RefCOCO-Gaze training set for a fair comparison. Then, similarly to the prior baselines, we train a speaker against this listener with \textsc{REC-Success} rewards. Once again, we find that exposure to RefCOCO does not result in higher accuracy or shorter references. 

\begin{table}[H]
\centering
\small
\begin{tabular}{lccc}
\toprule
\textbf{Speaker} & \textbf{Acc.} $\uparrow$ & \textbf{Length} & \textbf{$d_{\pi^s}$} $\downarrow$ \\
\midrule
\multicolumn{4}{c}{\textit{Overall (Averaged)}} \\
\midrule
\textsc{Molmo}                 & 45.65           & 16.35          & 1.41 \\
\textsc{REC-BeforeFirstHit}     & 46.18           & \phantom{0}3.05 & 0.99 \\
\textsc{REC-SeqAnyHit}         & 57.96           & 25.29          & 1.84 \\
\textsc{REC-Success}           & 61.94           & 19.70          & 1.41 \\
\textsc{REC-Shaping}           & 52.12           & \phantom{0}5.17 & 0.86 \\
\textsc{Gaze-SeqAnyHit}        & 51.78           & \phantom{0}8.87 & 0.95 \\
\textsc{Gaze-SeqLPHit}         & 54.25           & 11.90          & 1.03 \\
\textsc{Gaze-Shaping}          & 51.36           & \phantom{0}4.88 & 0.87 \\
\textsc{Gaze-BeforeFirstHit}   & 54.20           & \phantom{0}5.15 & 0.81 \\
\textsc{Human (Gold)}          & 88.98           & \phantom{0}3.52 & 0.00 \\
\midrule
\multicolumn{4}{c}{\textit{Baseline: RefCOCO-Finetuned Listener}} \\
\midrule
\textsc{REC-Success (RefCOCO Listener)} & 59.93 & 21.64 & 1.57 \\
\bottomrule
\end{tabular}
\caption{Performance comparison when training the speaker against a listener finetuned directly on RefCOCO.}
\label{tab:refcoco_finetuned_listener}
\end{table}

\subsection{Comparison to SFT}
To establish an upper bound, we finetune Molmo-7B-D directly on the RefCOCO validation set (8.8k examples). We consider 1\%, 10\%, and 100\% of this train set. Each model is trained for 4 epochs at learning rate 5e-5. As expected, direct supervision on human-written references yields strong performance. We additionally try training from the RefCOCO 1\% checkpoint with the \textsc{Gaze-Shaping} and \textsc{Gaze-SeqAnyHit} reward setups; however, this degrades performance from the SFT checkpoint.

\begin{table}[H]
\centering
\small
\begin{tabular}{lccc}
\toprule
\textbf{Speaker} & \textbf{Acc.} $\uparrow$ & \textbf{Length} & \textbf{$d_{\pi^s}$} $\downarrow$ \\
\midrule
\multicolumn{4}{c}{\textit{Overall (Averaged)}} \\
\midrule
\textsc{Molmo}                 & 45.65           & 16.35          & 1.41 \\
\textsc{REC-BeforeFirstHit}     & 46.18           & \phantom{0}3.05 & 0.99 \\
\textsc{REC-SeqAnyHit}         & 57.96           & 25.29          & 1.84 \\
\textsc{REC-Success}           & 61.94           & 19.70          & 1.41 \\
\textsc{REC-Shaping}           & 52.12           & \phantom{0}5.17 & 0.86 \\
\textsc{Gaze-SeqAnyHit}        & 51.78           & \phantom{0}8.87 & 0.95 \\
\textsc{Gaze-SeqLPHit}         & 54.25           & 11.90          & 1.03 \\
\textsc{Gaze-Shaping}          & 51.36           & \phantom{0}4.88 & 0.87 \\
\textsc{Gaze-BeforeFirstHit}   & 54.20           & \phantom{0}5.15 & 0.81 \\
\textsc{Human (Gold)}          & 88.98           & \phantom{0}3.52 & 0.00 \\
\midrule
\multicolumn{4}{c}{\textit{Comparison to SFT Baselines}} \\
\midrule
\textsc{RefCOCO SFT 100\%}               & 64.44 & \phantom{0}3.34 & 0.59 \\
\textsc{RefCOCO SFT 10\%}                & 64.56 & \phantom{0}3.17 & 0.59 \\
\textsc{RefCOCO SFT 1\%}                 & 60.27 & \phantom{0}3.28 & 0.71 \\
\textsc{RefCOCO SFT 1\% + Shaping Gaze}  & 55.21 & \phantom{0}3.79 & 0.84 \\
\textsc{RefCOCO SFT 1\% + Binary Gaze}   & 59.54 & \phantom{0}7.34 & 0.78 \\
\bottomrule
\end{tabular}
\caption{Comparison of Supervised Finetuning (SFT) baselines and SFT+RL hybrid models against the default RL/Gaze speakers.}
\label{tab:sft_comparison}
\end{table}

\subsection{Generalization to other VLMs}
We recreate a subset of the main results, instead training PaliGemma-3B-pt-448 \citep{beyer2024paligemmaversatile3bvlm} and LLaVA-1.5-7B \citep{liu2024improvedbaselinesvisualinstruction} as speaker models. This subset includes \textsc{Gaze-Shaping}, \textsc{Gaze-BeforeFirstHit}, \textsc{Gaze-SeqAnyHit}, \textsc{REC-Shaping}, \textsc{REC-BeforeFirstHit}, \textsc{REC-SeqAnyHit}, and \textsc{REC-Success}.

Table~\ref{tab:vlm_generalization} reports results; we observe that gains do not cleanly generalize across VLMs. Because hyperparameters were held constant from the Molmo-7B experiments, we attribute this to a lack of model-specific tuning, especially for the KL penalty coefficient ($\beta$), and differences in base models and their initial policy entropy.
\begin{table}[H]
\centering
\small
\begin{tabular}{lccc}
\toprule
\textbf{Speaker} & \textbf{Acc.} $\uparrow$ & \textbf{Length} & \textbf{$d_{\pi^s}$} $\downarrow$ \\
\midrule
\textsc{Human (Gold)}            & 88.98 & \phantom{0}3.52 & 0.00 \\
\multicolumn{4}{c}{\textit{Molmo-7B}} \\
\midrule
\textsc{Molmo}                  & 45.65 & 16.35           & 1.41 \\
\textsc{REC-BeforeFirstHit}     & 46.18 & \phantom{0}3.05 & 0.99 \\
\textsc{REC-SeqAnyHit}         & 57.96 & 25.29           & 1.84 \\
\textsc{REC-Success}           & 61.94 & 19.70           & 1.41 \\
\textsc{REC-Shaping}           & 52.12 & \phantom{0}5.17 & 0.86 \\
\textsc{Gaze-SeqAnyHit}        & 51.78 & \phantom{0}8.87 & 0.95 \\
\textsc{Gaze-SeqLPHit}         & 54.25 & 11.90           & 1.03 \\
\textsc{Gaze-Shaping}          & 51.36 & \phantom{0}4.88 & 0.87 \\
\textsc{Gaze-BeforeFirstHit}   & 54.20 & \phantom{0}5.15 & 0.81 \\
\midrule
\multicolumn{4}{c}{\textit{LLaVA-1.5-7B}} \\
\midrule
\textsc{LLaVA-1.5}              & 28.32 & \phantom{0}6.16 & 1.59 \\
\textsc{REC-BeforeFirstHit}     & 37.08 & \phantom{0}1.92 & 1.34 \\
\textsc{REC-Shaping}           & 44.40 & \phantom{0}2.04 & 1.14 \\
\textsc{REC-Success}           & 53.03 & 12.06           & 1.12 \\
\textsc{Gaze-BeforeFirstHit}   & 37.08 & \phantom{0}1.92 & 1.31 \\
\textsc{Gaze-Shaping}          & 20.51 & \phantom{0}3.19 & 1.79 \\
\textsc{Gaze-SeqAnyHit}        & 39.48 & \phantom{0}3.02 & 1.27 \\
\midrule
\multicolumn{4}{c}{\textit{PaliGemma-3B}} \\
\midrule
\textsc{PaliGemma}              & 31.80 & \phantom{0}3.08 & 1.48 \\
\textsc{REC-BeforeFirstHit}     & 41.29 & \phantom{0}0.96 & 1.24 \\
\textsc{REC-Shaping}           & 41.08 & \phantom{0}0.96 & 1.24 \\
\textsc{REC-SeqAnyHit}         & 34.17 & 26.09           & 2.28 \\
\textsc{REC-Success}           & 42.92 & \phantom{0}1.55 & 1.19 \\
\textsc{Gaze-BeforeFirstHit}   & 32.84 & \phantom{0}1.05 & 1.46 \\
\textsc{Gaze-Shaping}          & 29.38 & \phantom{0}0.92 & 1.56 \\
\textsc{Gaze-SeqAnyHit}        & 35.11 & \phantom{0}8.00 & 1.43 \\
\bottomrule
\end{tabular}
\caption{Cross-architecture evaluation using LLaVA-1.5-7B and PaliGemma-3B as speaker backbones.}
\label{tab:vlm_generalization}
\end{table}

\section{Quality and qualitative analyses} \label{sec:qualitative_error_analysis}

\subsection{Strategy analysis} \label{appx:refstrategy}
To consider not only the similarity of length and accuracy to human-generated references, but also stylistic choices, we perform a simple analysis of referring expression strategy. Referring expressions are often separated into spatial or attribute based references; using keyword lists, we show the proportion of references that use spatial or attribute based keywords, both, or neither. The keywords used for each category were:
\begin{itemize}
    \item \textbf{Spatial Keywords:} left, right, top, bottom, middle, center, above, below, next, near, far, between, behind, front, centered, side, corner, upper, lower, background, foreground, back, overhead, touching, overlapping, contacting, abutting, alongside, bordering, nearest, farthest, furthest, closest, distant, adjacent, beside, inside, outside, within, interior, exterior, embedded, surrounded, among, amongst, amid, under, underneath, beneath, atop, upon, over, edge, margin, boundary, perimeter, diagonal, aligned, stacked, facing, opposite, rear, first, second, third, last, outer, inner, innermost, outermost
    
    \item \textbf{Attribute Keywords:} red, blue, green, yellow, orange, purple, pink, brown, black, white, gray, grey, silver, gold, tan, beige, maroon, navy, teal, cyan, magenta, violet, dark, light, bright, pale, deep, vivid, transparent, translucent, opaque, multicolored, colorful, neon, turquoise, indigo, cream, charcoal, blonde, brunette, auburn, large, small, big, little, tall, short, long, round, circular, square, rectangular, triangular, oval, spherical, cylindrical, flat, curved, pointed, straight, crooked, hollow, solid, narrow, wide, thick, thin, slender, heavy, metal, metallic, wooden, plastic, glass, leather, denim, cotton, woolen, ceramic, stone, brick, concrete, paper, smooth, rough, soft, hard, fluffy, hairy, furry, shiny, glossy, matte, rusty, wet, dry, striped, checkered, patterned, spotted, dotted, printed, wearing, holding, carrying, sitting, standing, walking, running, smiling, lying, reclining, leaning, crouching, kneeling, jumping, riding, flying, swimming, eating, drinking, looking, pointing, reading, sleeping, driving, hanging, attached, mounted, folded, elderly, young, old, new, broken, dirty, clean, damaged, stained, burnt, full, empty, open, closed, bare, decorated, painted, covered, single, double, pair, couple, group, crowd, several, few, many
\end{itemize}

\begin{figure}[H]
\begin{center}
    \includegraphics[width=\linewidth]{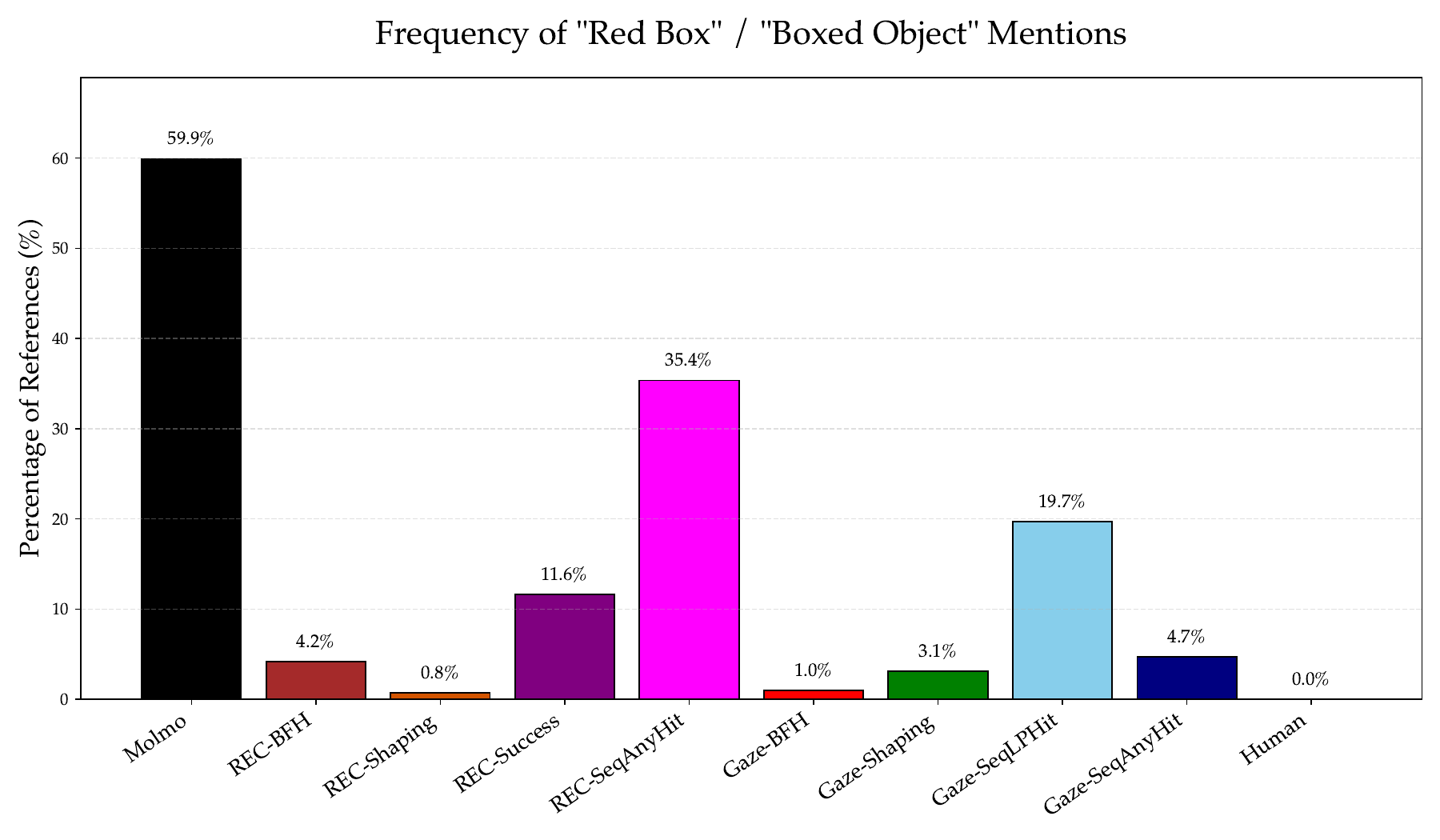}
\end{center}
\caption{Mention frequency of the red bounding box across speaker models.}
\label{fig:redbox}
\end{figure}

A qualitative observation of references produced by Molmo without any additional training was that many of them started by restating the prompt by mentioning the target region identifier --- in this case, the "red box". This is a clear indicator that the models are not pragmatic, as they can not distinguish whether this information is useful for a listener. We investigate which reward setups successfully decrease the usage of such prefixes in Fig~\ref{fig:redbox}.

\subsection{Reference quality annotations} \label{subsec:refquality}
We perform a small scale analysis of the quality and ambiguity of references. On a subset of reward configurations, one author annotated a subset of 50 image-reference pairs, sampled from each dataset as either well-specified, ambiguous, or incorrect. Configurations that encourage longer references such as \textsc{REC-Success} and \textsc{Gaze-SeqAnyHit} were more likely to be well-specified. \textsc{Gaze-Shaping} encourages shorter references, but also produces more ambiguous references, demonstrating the tradeoff between accuracy and efficiency. 

\begin{figure}[H]
\begin{center}
    \includegraphics[width=\linewidth]{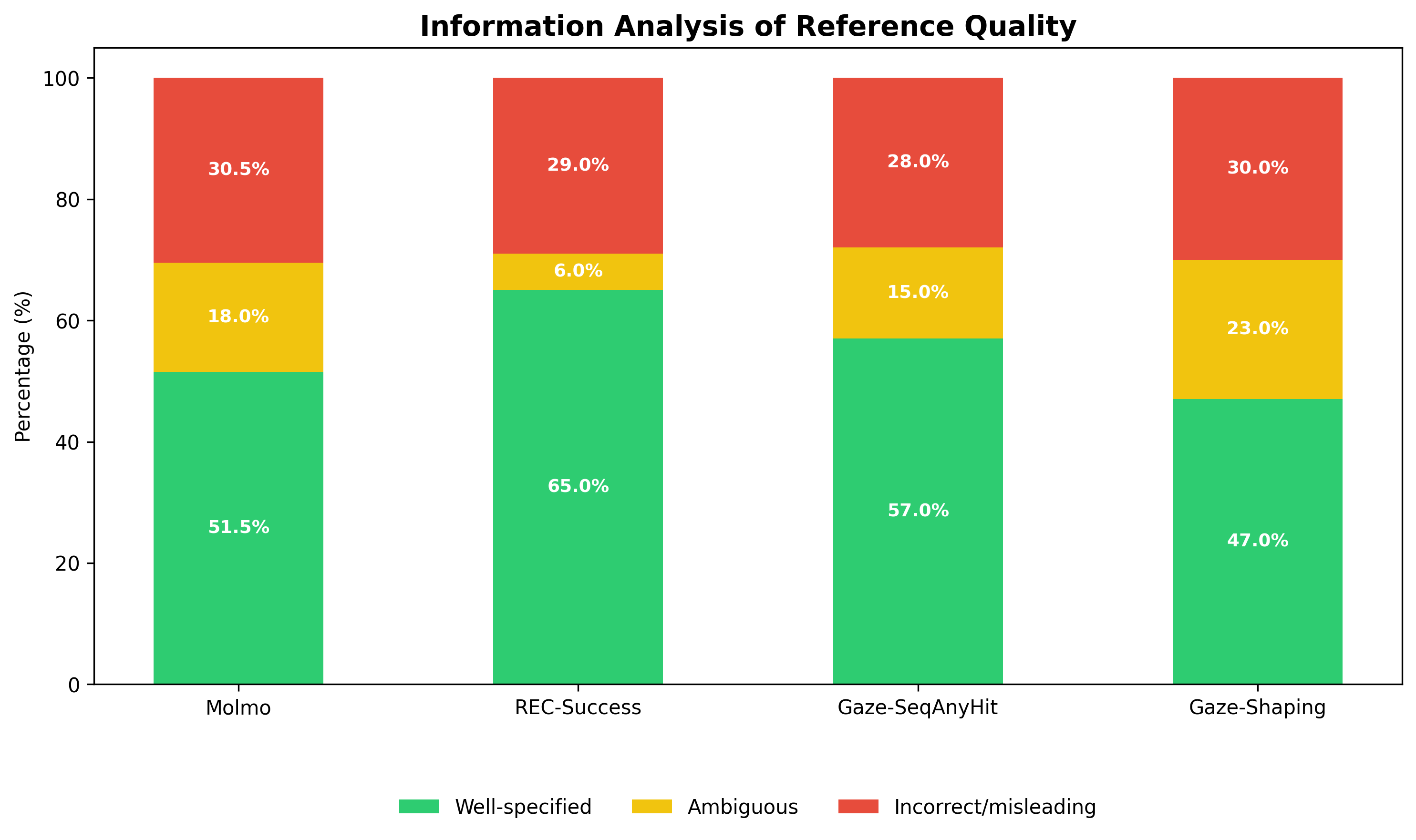}
\end{center}
\caption{Analysis of ambiguity and incorrect reference rates across four selected speaker models.}
\end{figure}

\subsection{Qualitative behavior and error discussion}
\begin{figure}[h]
\begin{center}
    \includegraphics[width=\linewidth]{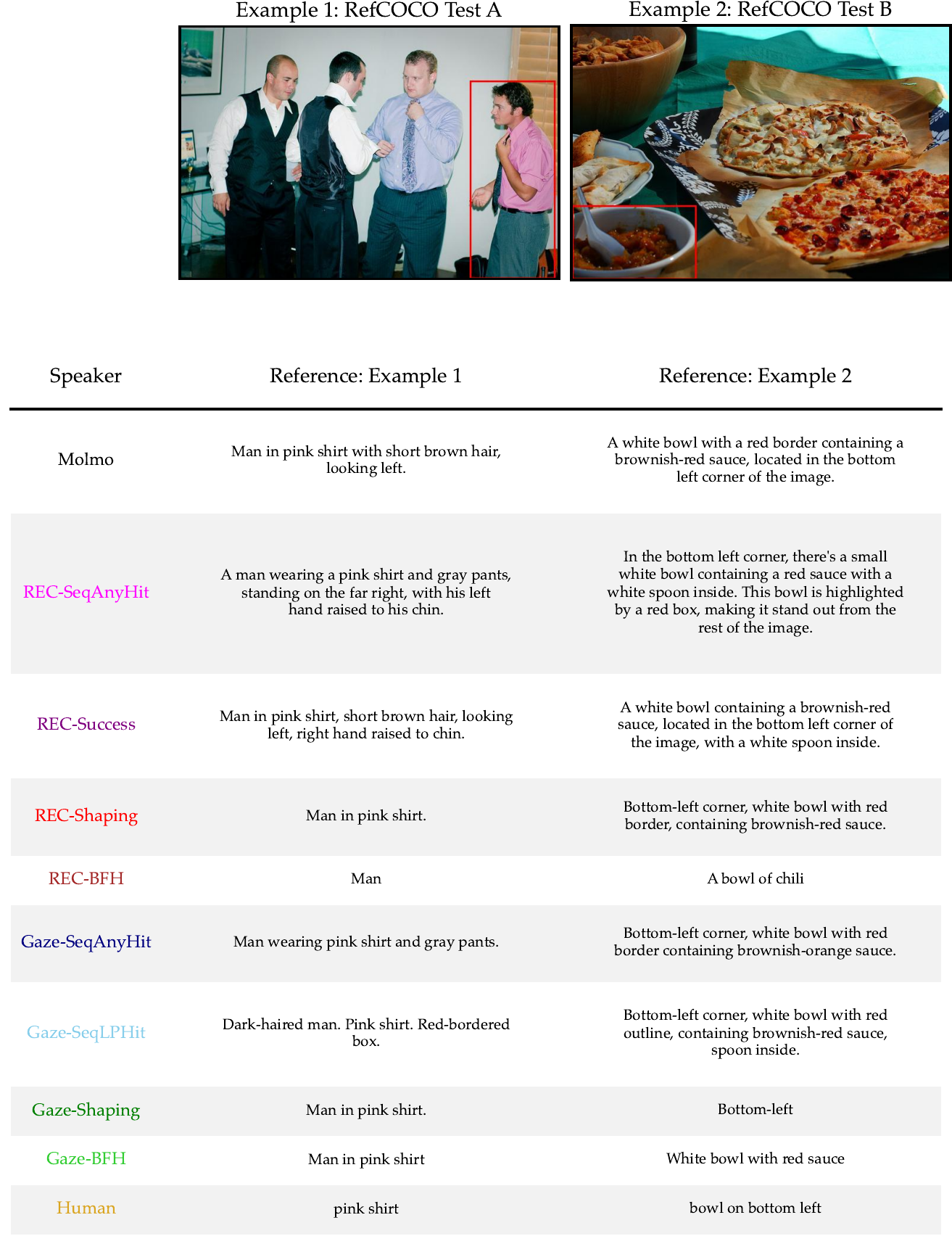}
\end{center}
\caption{Examples of generated references from all speaker models.}
\label{appx:examples}
\end{figure}

\subsection{Gaze retention analysis}
 \textsc{Gaze-BFH} is the best performing model with machine listener, but it falls short with human listeners. Thus, we investigate why this might be the case. While not displayed in the examples in Fig~\ref{appx:examples}, we suspect the \textsc{Gaze-BFH} configuration underperforms with human listeners due to its tendency to reward false positives. To corroborate this hypothesis, we analyze the post-hit target retention rate across speaker models, defined as the conditional probability that a gaze trajectory (or sequence of REC points) remains within the target bounding box after initial entry. Analysis results are shown in Table~\ref{tab:gaze_trajectory_behavior}. Specifically, we evaluate the generated point sequences from the validation episodes of each model's best checkpoint (the checkpoint selected for evaluation). Overall, the gaze predictor is more prone to exiting the target region than the REC model (Molmo) since Molmo is trained for REC rather than realistic scanpaths. \textsc{Gaze-BFH} displays the worst retention rate of all speaker models. The reward setup does not encourage references that keep the listener's gaze in the bounding box. \textsc{Gaze-SeqAnyHit} also displays very poor retention rate, but likely does not suffer during human evaluation due to longer generations including more information.

\begin{table}[htbp]
\centering
\small
\begin{tabular}{lc}
\toprule
\textbf{Speaker Model} & \textbf{Gaze Retention Rate (\%)}  \\
\midrule
\textsc{REC-Success}    & 100.0\% \\
\textsc{REC-BFH}        &  90.6\% \\
\textsc{REC-Shaping}    &  90.4\% \\
\textsc{REC-SeqAnyHit}  &  85.7\% \\
\textsc{Gaze-Shaping}   &  54.3\% \\
\textsc{Gaze-SeqLPHit}  &  52.8\% \\
\textsc{Gaze-SeqAnyHit} &  39.9\% \\
\textsc{Gaze-BFH}        &  37.2\% \\
\bottomrule
\end{tabular}
\caption{Gaze trajectory retention metrics across speaker model configurations, demonstrating post-hit retention across reward setups.}
\label{tab:gaze_trajectory_behavior}
\end{table}

\subsection{Punctuation and spacing errors}
We observe that some of the configurations result in list-like references of many short sentences (e.g., \textsc{Gaze-SeqLPHit} in Fig~\ref{appx:examples}). Additionally, \textsc{REC-Shaping} inserts extra spaces. To quantify these phenomena, we record the double-space rate of each speaker model along with the average number of periods in each reference. The latter metric is based on the assumption that most references are single sentences or phrases. In Table~\ref{tab:text_formatting_artifacts}, \textsc{Gaze-SeqLPHit}, \textsc{Gaze-SeqAnyHit}, and \textsc{REC-SeqAnyHit} are particularly prone to inserting extra periods. Only \textsc{REC-Shaping} develops the double space behavior; sometimes, the speaker generates spaces between tokens that make up a single word, affecting the speech synthesis during human evaluation. This exemplifies how the human evaluation accounts for behaviors that are otherwise difficult to quantify. We hypothesize both of these behaviors develop to exploit the listener model by giving it extra chances to resolve the reference through adding extra tokens without any additional semantic value. 

\begin{table}[htbp]
\centering
\small
\begin{tabular}{lccc}
\toprule
\textbf{Speaker Model} & \textbf{Mid-Ref Period Rate (\%)} $\downarrow$ & \textbf{Double Space Rate (\%)} $\downarrow$ & \textbf{Mean Periods} \\
\midrule
\textsc{Molmo}                 & 18.58 & \phantom{0}0.00 & 1.22 \\
\textsc{REC-BeforeFirstHit}    & \phantom{0}0.78 & \phantom{0}0.00 & 0.02 \\
\textsc{REC-Shaping}           & \phantom{0}6.93 & 53.17 & 0.84 \\
\textsc{REC-Success}           & 17.12 & \phantom{0}0.00 & 1.20 \\
\textsc{REC-SeqAnyHit}         & 39.31 & \phantom{0}0.00 & 1.46 \\
\textsc{Gaze-BeforeFirstHit}   & \phantom{0}0.52 & \phantom{0}0.00 & 0.04 \\
\textsc{Gaze-Shaping}          & \phantom{0}4.67 & \phantom{0}0.00 & 0.42 \\
\textsc{Gaze-SeqLPHit}        & 53.54 & \phantom{0}0.00 & 2.87 \\
\textsc{Gaze-SeqAnyHit}        & 28.57 & \phantom{0}0.00 & 1.82 \\
\textsc{Human}                 & \phantom{0}0.46 & \phantom{0}0.32 & 0.02 \\
\bottomrule
\end{tabular}
\caption{Frequency of mid-reference period punctuation, double spacing, and mean period counts across speaker model configurations.}
\label{tab:text_formatting_artifacts}
\end{table}

\subsection{Coordinate generation}
Finally, we occasionally observe references that are simply coordinates in \textsc{REC-BFH} (Table~\ref{tab:coordinate_occurrence_rate}). Although not incredibly prevalent, this case highlights how using non human-like listeners can result in reward-hacking by creating references that exploit the machine listener. 

\begin{table}[htbp]
\centering
\small
\begin{tabular}{lc}
\toprule
\textbf{Speaker Model} & \textbf{Coordinate Occurrence Rate (\%)} $\downarrow$ \\
\midrule
\textsc{Molmo}                 & 0.00\% \\
\textsc{REC-BeforeFirstHit}    & \textbf{0.78\%} \\
\textsc{REC-Shaping}           & 0.02\% \\
\textsc{REC-Success}           & 0.00\% \\
\textsc{REC-SeqAnyHit}         & 0.01\% \\
\textsc{Gaze-BeforeFirstHit}   & 0.00\% \\
\textsc{Gaze-Shaping}          & 0.03\% \\
\textsc{Gaze-SeqLPHit}        & 0.01\% \\
\textsc{Gaze-SeqAnyHit}        & 0.01\% \\
\textsc{Human}                 & 0.00\% \\
\bottomrule
\end{tabular}
\caption{Frequency of coordinate artifact occurrences across speaker model outputs.}
\label{tab:coordinate_occurrence_rate}
\end{table}

\section{Human evaluation details} \label{humanevaldetails}

We recruit 751 participants on Prolific who are fluent English speakers in the United States. Participants receive \$2 USD per session, each of which takes an average of 6.1 minutes. In each session, a participant completes up to 40 trials, with each trial corresponding to an image paired with a referring expression. We record all mouse movements, including clicks; only after the audio has finished playing will a click on the image move the participant to the next trial. Five of the 40 trials include human-written referring expressions; participants are provided a bonus of \$0.50 USD if they successfully identify the target referent for all 5 human-written references without revealing the reference text. This bonus structure both incentivizes participants to perform well as a listener, and discourages them from revealing the reference text when unnecessary. To reduce spam and low-effort sessions, we discard all data from participants who do not successfully identify at least 4 of the 5 human-written references, resulting in a final pool of 633 participants.

\subsection{Experimental instructions}
The following instructions were provided to Prolific participants prior to beginning the reference game:

\begin{table}[H]
\begin{tcolorbox}[colback=white, colframe=black!70, arc=0pt, outer arc=0pt, title=Experimental Instructions (Verbatim)]
\small
\textbf{Welcome!} This task consists of 40 or less trials. When each trial begins, you will see an image, and hear a description read aloud. Your job is to click on the region of the image that best matches the description.

\vspace{0.5em}
\textbf{Bonus!} We will check your answers on a subset of trials with unambiguous descriptions. Participants who get 100\% on this subset will receive a \$0.50 bonus.

\vspace{0.5em}
\textbf{Need to see the description?} If the speech description is unclear, you have the option to reveal the description text with a button below the image. This button will only appear after the speech description finishes. \textbf{However}, if you reveal the description for examples from the unambiguous subset, you will not receive the bonus, so only reveal it when you truly need it.

\vspace{0.5em}
\textbf{What if you don't know?} There will be some ambiguous descriptions. If you are unsure, just make your best guess! 

\vspace{0.5em}
\textbf{Ok, ready to begin?} Please make sure your volume is turned up. The task will begin on the next page.
\end{tcolorbox}
\end{table}

\end{document}